%% file: SAMA.tex
\documentclass[UTF8]{article}

    \PassOptionsToPackage{numbers, compress}{natbib}

\usepackage[preprint]{SAMA}

\usepackage[utf8]{inputenc} 
\usepackage[T1]{fontenc}    
\usepackage{hyperref}       
\usepackage{url}            
\usepackage{wrapfig,lipsum}
\usepackage{booktabs}       
\usepackage{amsfonts}       
\usepackage{nicefrac}       
\usepackage{microtype}      
\usepackage{xcolor}         
\usepackage{booktabs}

\usepackage{caption}
\usepackage{subcaption}
\usepackage{bbding}
\usepackage{graphicx}
\usepackage{multirow}

\newcommand{\eg}{\textit{e}.\textit{g}.}

\AtEndPreamble{
    \usepackage[capitalize]{cleveref}
    \crefname{section}{Sec.}{Secs.}
    \Crefname{section}{Section}{Sections}
    \Crefname{table}{Table}{Tables}
    \crefname{table}{Tab.}{Tabs.}
}

\usepackage{colortbl}
\definecolor{mygray}{gray}{0.95}
\definecolor{my_green}{RGB}{82,208,80}
\usepackage{enumerate}
\usepackage{makecell}
\definecolor{00red}{RGB}{236,35,35}

\usepackage{pifont}

\definecolor{mygreen}{RGB}{112, 180, 143}
\definecolor{myred}{RGB}{242, 128, 128}

\definecolor{lightgray}{gray}{.9}
\definecolor{lightblue}{RGB}{230,240,255}
\definecolor{lightgreen}{RGB}{230,255,230}
\definecolor{lightyellow}{RGB}{255,255,230}
\definecolor{lightred}{RGB}{255,230,230}

\definecolor{lightlightgray}{gray}{.95}
\definecolor{lightlightblue}{RGB}{240,245,255}
\definecolor{lightlightgreen}{RGB}{240,255,240}
\definecolor{lightlightyellow}{RGB}{255,255,240}
\definecolor{lightlightred}{RGB}{255,240,240}

\definecolor{lightlightlightgray}{gray}{.99}
\definecolor{lightlightlightblue}{RGB}{247,250,255}
\definecolor{lightlightlightgreen}{RGB}{247,255,247}
\definecolor{lightlightlightyellow}{RGB}{255,255,247}
\definecolor{lightlightlightred}{RGB}{255,247,247}

\usepackage[normalem]{ulem}
\usepackage{comment}

\usepackage{amsmath}
\usepackage{fancyhdr}

\usepackage{tabularx}
\usepackage{adjustbox}

\usepackage[most]{tcolorbox}
\definecolor{BoxBackground}{RGB}{240, 240, 240} 
\definecolor{BoxFrame}{RGB}{0, 0, 0} 
\definecolor{TitleBackground}{RGB}{0, 0, 0} 
\definecolor{TitleText}{RGB}{255, 255, 255} 

\usepackage{makecell}
\usepackage{multirow}
\usepackage[table]{xcolor}
\usepackage{float}

\definecolor{oursrow}{RGB}{230,245,255}
\definecolor{BoxBackground}{RGB}{240, 240, 240} 
\definecolor{lightlightgray}{gray}{.95}

\usepackage{pifont}
\usepackage{tabularx}
\usepackage{adjustbox}
\usepackage{wrapfig}

\usepackage[most]{tcolorbox}   
\usepackage{enumitem}          
\newtcolorbox{AcademicBox}[1]{%
  colback=gray!3,
  colframe=black!35,
  boxrule=0.4pt,
  arc=2pt,
  left=6pt,right=6pt,top=5pt,bottom=5pt,
  width=0.92\linewidth,
  title={#1}
}

\tcbset{
  academicbox/.style={
    boxsep=5pt,
    left=2pt,
    right=2pt,
    bottom=0.5pt,
    boxrule=0.5pt,
    colback=BoxBackground,
    colframe=BoxFrame,
    colbacktitle=TitleBackground,
    coltitle=TitleText,
    enhanced,
    attach boxed title to top left={yshift=-0.1in,xshift=0.1in},
    boxed title style={boxrule=0pt,colframe=white},
    title={#1},
  }
}

\usepackage{booktabs}
\usepackage{multirow}
\usepackage{makecell}
\usepackage{graphicx}
\usepackage{amssymb}

\title{SAMA: Factorized Semantic Anchoring and Motion Alignment for Instruction-Guided Video Editing}
\author{%
  Xinyao Zhang$^{1,2}\textsuperscript{*}$,
  Wenkai Dong$^{1}\textsuperscript{*}$, 
  Yuxin Song$^{1}\textsuperscript{*$\dagger$}$, 
  Bo Fang$^{1,3}$, Qi Zhang$^1$, Jing Wang$^{1,2}$, \\
  \textbf{Fan Chen$^1$, Hui Zhang$^1$, Haocheng Feng$^1$, Yu Lu$^{4}\textsuperscript{$\ddagger$}$, Hang Zhou$^1$, Chun Yuan$^2$, Jingdong Wang$^1$} \\
  $^1$Baidu \quad
  $^2$Tsinghua University \quad  
  $^3$City University of Hong Kong \quad
  $^4$Zhejiang University
  \\
  \tt\small {Project Page: \url{https://cynthiazxy123.github.io/SAMA}} \\
  \tt\small {Email Address: songyuxinbb@outlook.com} \\
}

\begin{document}
\maketitle
\begingroup
\renewcommand\thefootnote{}
\footnotetext{$^{*}$Equal Contribution \quad 
$^{\dagger}$Project Leader \quad 
$^{\ddagger}$Corresponding Author}
\endgroup
\vspace{-5mm}


\input{sec/0_abstract}

\begin{figure}[!t]
  \centering
  \includegraphics[width=\textwidth]{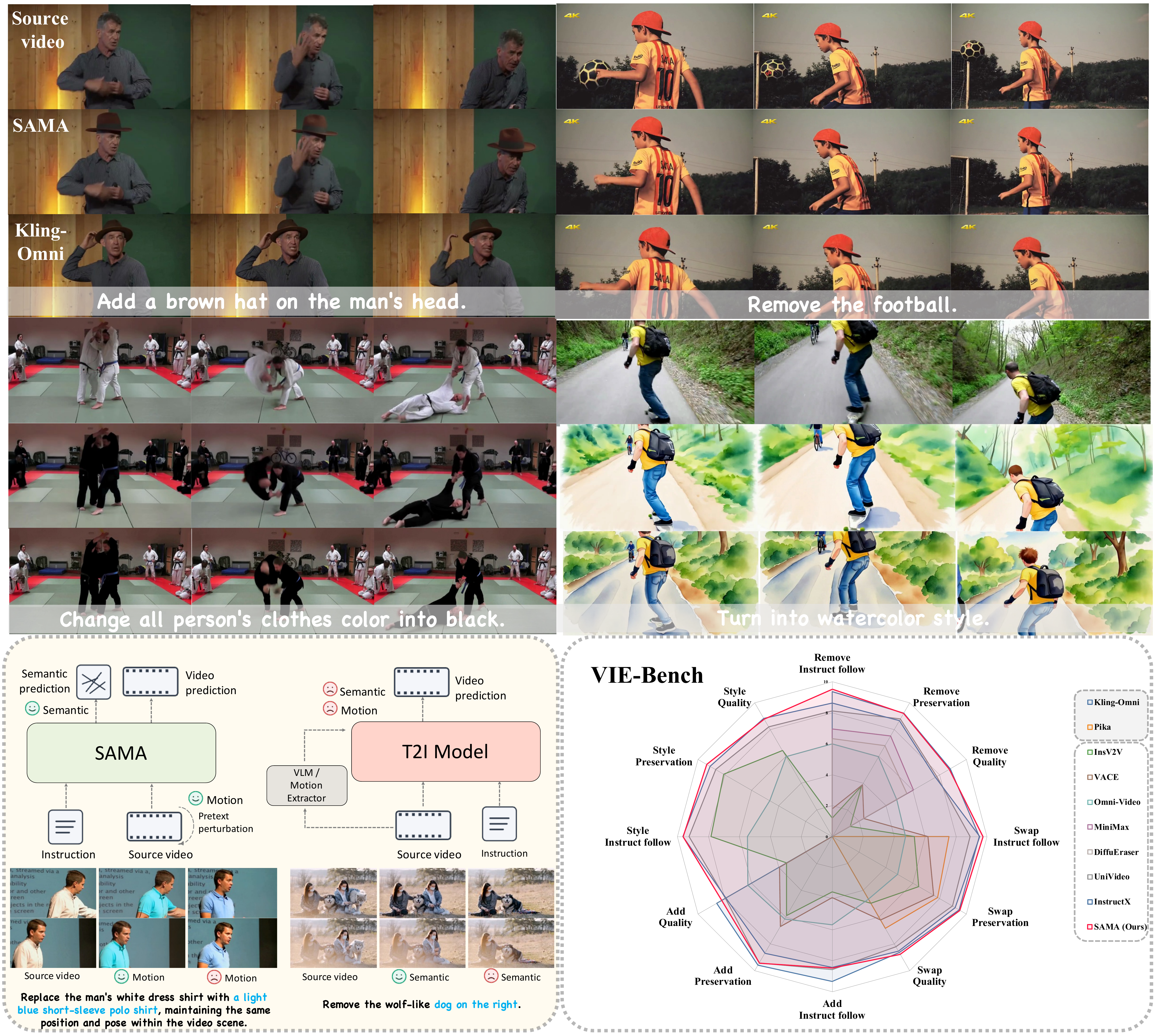}
  \caption{\textbf{Teaser and overview.} Top: qualitative comparisons on VIE-Bench, comparing SAMA with representative open- and closed-source systems. Bottom left: illustration of SAMA’s semantic--motion training objectives. Bottom right: fine-grained VIE-Bench performance comparison.}
  \label{fig:intro_teaser}
\end{figure}

\input{sec/1_intro}
\input{sec/2_related}
\input{sec/3_method}
\input{sec/4_exp}

\input{sec/5_conclusion}

\clearpage
\bibliographystyle{unsrt}

\bibliography{SAMA}

\clearpage
\appendix
\input{sec/7_appendix}

\end{document}

%% file: sec/0_abstract.tex
\begin{abstract}
Current instruction-guided video editing models struggle to simultaneously balance precise semantic modifications with faithful motion preservation. While existing approaches rely on injecting explicit external priors (\eg, VLM features or structural conditions) to mitigate these issues, this reliance severely bottlenecks model robustness and generalization.
To overcome this limitation, we present \textbf{SAMA} (factorized \textbf{S}emantic \textbf{A}nchoring and \textbf{M}otion \textbf{A}lignment), a framework that factorize video editing into semantic anchoring and motion modeling.
First, we introduce \emph{Semantic Anchoring} which establish a reliable visual anchor by jointly predicting semantic tokens and video latents at sparse anchor frames, enabling purely instruction-aware structural planning.
Second, \emph{Motion Alignment} pre-trains the same backbone on motion-centric video restoration pretext tasks (cube inpainting, speed perturbation, and tube shuffle), enabling the model to internalize temporal dynamics directly from raw videos.
SAMA is optimized with a two-stage pipeline: a factorized pre-training stage that learns inherent semantic-motion representations without paired video-instruction editing data, followed by supervised fine-tuning on paired editing data.
Remarkably, the factorized pre-training alone already yields strong \emph{zero-shot} video editing ability, validating the proposed factorization.
SAMA achieves state-of-the-art performance among open-source models and is competitive with leading commercial systems (e.g. Kling-Omni). Code, models, and datasets will be released.


\end{abstract}

%% file: sec/1_intro.tex
\section{Introduction}
\label{sec:intro}
Diffusion models have enabled interactive, instruction-guided image editing with impressive fidelity and controllability~\cite{instructpix2pix, magicbrush, ultraedit, anyedit, step1xedit, song2026cologen, seed-data-edit, wang2025seededit, wu2025qwenimage}. 
Extending this paradigm from single images to videos, however, remains substantially more challenging.
A practical instruction-guided video editor must \emph{(i)} apply fine-grained semantic changes that follow the instruction, while \emph{(ii)} preserving temporally coherent motion of the edited subject, background, and camera.
In current models, these two requirements often conflict: aggressive semantic changes induce localized artifacts, identity drift, and texture popping, whereas enforcing temporal consistency can dilute the intended edit and reduce instruction fidelity (Fig.~\ref{fig:intro_teaser} top). This tension has been widely observed in diffusion-based video editing and adaptation works~\cite{tuneavideo, tokenflow, videop2p, fatezero}.

To mitigate these issues, a prevailing trend in existing approaches is to rely on injecting explicit \textit{external} priors, such as VLM-extracted semantic conditions~\cite{instructx, klingomni} or structural signals like skeletons and depth maps~\cite{ zhangcontrolvideo, chen2023control}. 
We argue that this over-reliance reflects a significant bottleneck, which constrains the diffusion backbone from learning \emph{inherent semantic-motion representations} 
for precise semantic editing and faithful motion alignment with the source video dynamics.
Instead, we attribute the core difficulty of instruction-guided video editing to the lack of \emph{factorization} between semantic structure planning and motion modeling~\cite{moviegen, sora, kong2024hunyuanvideo, agarwal2025cosmos, ha2018worldmodels}.
Semantic edits are typically sparse and temporally stable: a small number of anchor frames is often sufficient to determine the desired visual modification.
In contrast, motion coherence follows physical and temporal dynamics that can be learned from large-scale raw videos without explicit editing supervision.

Based on this observation, we propose \textbf{SAMA} (factorized \textbf{S}emantic \textbf{A}nchoring and \textbf{M}otion \textbf{A}lignment), a framework that encourages the model to learn semantic structure planning and motion modeling as two complementary capabilities.
First, we introduce \emph{Semantic Anchoring} which predicts semantic tokens together with video latents to support instruction-aware structural planning in the semantic space while retaining high-fidelity rendering in the latent space.
Second, \emph{Motion Alignment} strengthens temporal reasoning through motion-centric video restoration tasks, encouraging the backbone to internalize coherent temporal dynamics directly from raw videos.

To realize this factorized learning paradigm, we train SAMA with a two-stage strategy. 
In the first stage, a \emph{factorized pre-training} process encourages the model to internalize semantic anchoring and motion dynamics as two complementary capabilities, without requiring paired instruction-guided video editing data. 
Remarkably, we find that this stage alone already induces strong \emph{zero-shot} video editing behavior. 
This observation suggests that robust instruction-guided video editing can naturally emerge once a model learns to jointly reason about semantic intent and temporal dynamics.
In the subsequent \emph{supervised fine-tuning} stage, the model is trained on paired video editing datasets to resolve residual semantic--motion conflicts and improve visual fidelity. Consequently, SAMA achieves state-of-the-art performance among open-source models while delivering results comparable to leading commercial systems (e.g. Kling-Omni~\cite{klingomni}, Runway~\cite{runwayaleph}).

\begin{itemize}
    \item We propose a factorized perspective on instruction-guided video editing that separates semantic planning from motion modeling, reducing reliance on brittle external priors.

    \item We introduce \emph{Semantic Anchoring} and \emph{Motion Alignment} via motion-centric video restoration pre-training, enabling the diffusion backbone to internalize robust semantic and temporal representations.

    \item SAMA achieves state-of-the-art performance among open-source video editing models and is competitive with leading commercial systems. Code, models, and datasets will be publicly released.
\end{itemize}

%% file: sec/2_related.tex
\section{Related Work}
\subsection{Instruction-Guided Video Editing}
Instruction-guided video editing aims to edit an input video following a text instruction, with the key challenge of preserving temporal consistency. Early diffusion-based attempts~\cite{geyer2023tokenflow, fatezero, couairon2023videdit, yang2023rerender, shen2024explanatory, cong2024flatten, tuneavideo, videop2p} in instruction-guided video editing mainly follow zero-shot or one-/few-shot paradigms, where pretrained text-to-image diffusion models are repurposed for videos with additional temporal modeling to maintain consistency.

With the release of large-scale instruction-guided video editing datasets such as Se\~norita-2M~\cite{senorita-2m}, InsViE-1M~\cite{insvie}, Ditto-1M~\cite{ditto}, ReCo-Data~\cite{reco}, and OpenVE-3M~\cite{he2025openve}, recent research has shifted toward data-driven video editing models trained end-to-end.
Ditto~\cite{ditto} builds its large-scale synthetic data pipeline by combining a strong image editing model with an in-context video generation model, and then trains a model on Ditto-1M to improve instruction-guided and temporal consistency. 
OpenVE-3M~\cite{he2025openve} expands supervision across diverse editing categories, while ReCo-Data~\cite{reco} focuses on region-aware instruction editing to improve local controllability.

Several recent works \cite{unic, diffueraser,vace, lucyedit, insv2v, icve, editverse, videocof, reco} further explore unified and in-context formulations for video editing.
UNIC~\cite{unic} unifies different video editing tasks by converting the noisy video latents, source video tokens, and multi-modal condition tokens into a single sequence, so a Diffusion Transformer can learn editing behaviors in-context without task-specific adapters or DDIM inversion.
VACE~\cite{vace} explores a unified and controllable editing formulation that supports diverse edit operations, improving the generality and robustness of instruction-guided video editing.
ICVE~\cite{icve} proposes a low-cost pretraining strategy that uses unpaired video clips to learn general editing ability in-context, and then refines the model with a small amount of paired editing data.
EditVerse~\cite{editverse} proposes a unified framework for image/video generation and editing by representing text, images, and videos in a shared token space, enabling strong in-context editing and supporting data-driven training with large-scale benchmarks.
DiffuEraser~\cite{diffueraser} studies instruction-guided video object removal by integrating diffusion-based editing with temporal-consistent inpainting, aiming to erase targets while preserving coherent backgrounds across frames.
ReCo~\cite{reco} introduces a joint source-target video diffusion framework and applies region constraints to improve instruction-guided editing.
VideoCoF~\cite{videocof} introduces a Chain-of-Frames ``see--reason--edit'' formulation that predicts where/how to edit across frames before generation, improving instruction-to-region alignment and temporal consistency without requiring user-provided masks.

Beyond editing-centric models, unified video understanding and generation frameworks such as Omni-Video~\cite{tan2025omni}, InstructX~\cite{instructx}, UniVideo~\cite{univideo}, and VINO~\cite{chen2026vino} provide strong representations for video content and motion dynamics.

\subsection{Semantic Alignment on Image and Video Generation}
Recent progress in image and video generation also benefits from semantic alignment between generative models and strong pretrained encoders. 
In image generation, REPA~\cite{REPA} aligns intermediate denoising features with clean features from a pretrained image encoder, which stabilizes training and improves generation quality. 
Following REPA, several works study how to apply representation alignment more effectively, including end-to-end VAE--diffusion training (REPA-E~\cite{leng2025repae}), stage-wise scheduling to avoid late-stage degradation (HASTE~\cite{wang2025haste}), teacher-free self-alignment via self-distillation (SRA~\cite{jiang2025sra}).

Similar ideas have recently been extended to video generation.
SemanticGen~\cite{bai2025semanticgen} first predicts compact semantic features and then generates VAE latents conditioned on them, which is more efficient for long videos.
VideoREPA~\cite{zhang2025videorepa} distills spatio-temporal relational knowledge from video foundation models into text-to-video diffusion models via token-relation alignment.
Beyond generation, this relational alignment idea has been adopted for video editing: FFP-300K~\cite{huang2026ffp} uses inter-frame relational distillation inspired by VideoREPA to better preserve source motion.

\noindent\textbf{Positioning.}
Inspired by recent advances in semantic alignment for image/video generation, we apply semantic-alignment regularization to instruction-guided video editing. Our approach improves instruction following and temporal consistency, and accelerates DiT convergence during training, without heavy test-time optimization.

\subsection{Self-supervised Learning for Video Representation Learning}
Self-supervised learning learns spatiotemporal representations from unlabeled videos via pretext tasks. Motivated by this line of work, we adopt lightweight pretext tasks as motion-centric restoration objectives in our \textit{Motion Alignment} (Sec.~\ref{sec:ma}) to better capture coherent temporal dynamics. Prior works mainly fall into three categories: speed-based learning (e.g., SpeedNet~\cite{benaim2020speednet}, PRP~\cite{yao2020videoprp}, Pace Prediction~\cite{wang2020pace}), spatiotemporal puzzles (e.g., Space-Time Cubic Puzzles~\cite{kim2019cubicpuzzles}), and reconstruction-based objectives (e.g., masked video modeling and VideoMAE~\cite{tong2022videomae}).

%% file: sec/3_method.tex
\begin{figure}[t]
  \centering
  \includegraphics[width=0.95\linewidth]{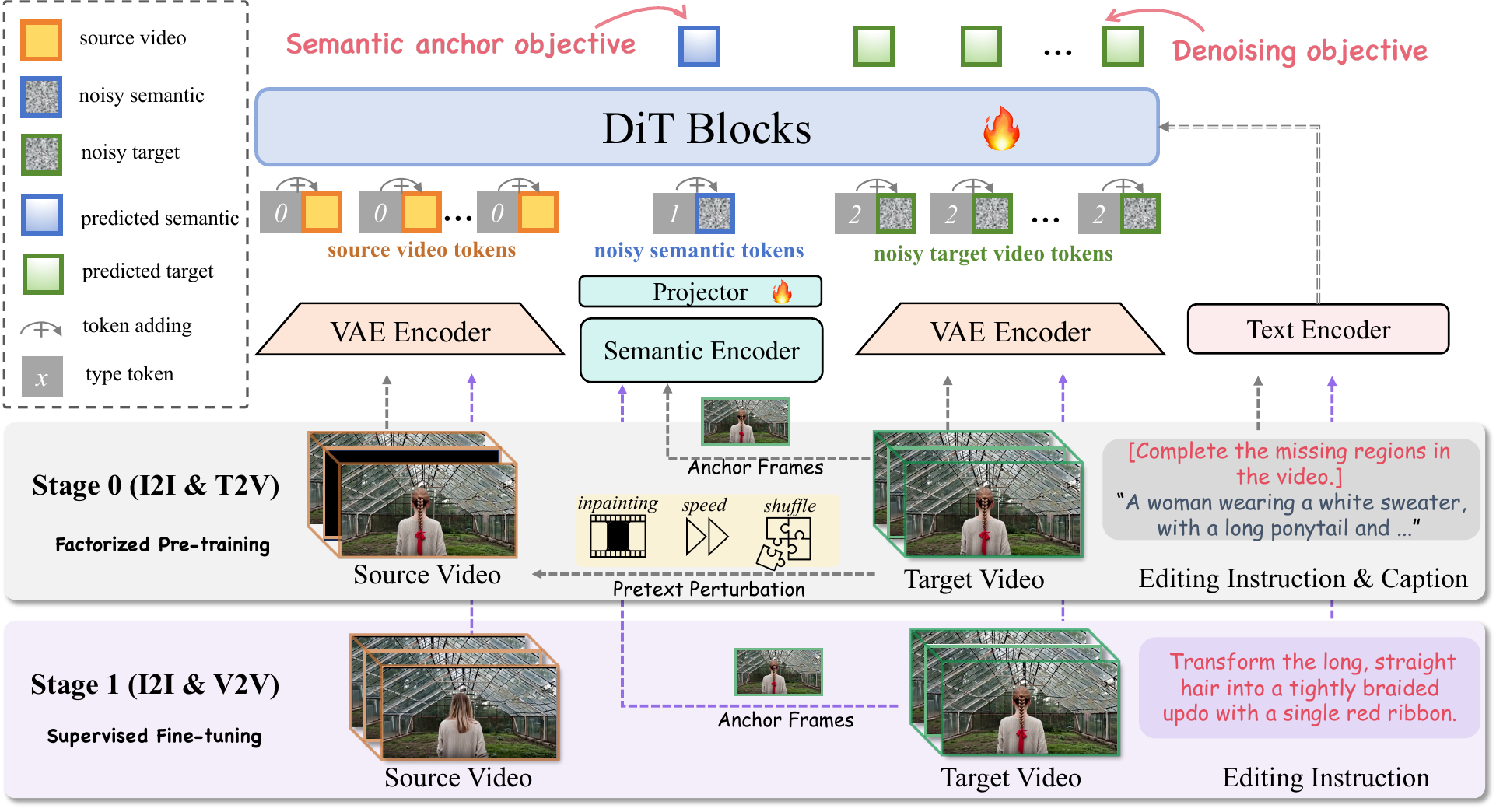} 
  \caption{\textbf{Overall pipeline.} 
  SAMA first performs factorized pre-training (stage 0) on additional perturbed videos by completing a pretext task conditioned on the given captions. 
  It then performs normal supervised fine-tuning (stage 1) on original source videos.
  Semantic Anchoring is incorporated in both stages to jointly facilitate semantic representation learning and instruction-guided video editing.
   }
  \label{fig:pipeline}
\end{figure}

\section{Method}
\textbf{Preliminary} We adopt a video diffusion transformer framework trained via the flow matching~\cite{flowmatching} paradigm. The main training objective is to minimize the expected flow matching loss, defined as:
\begin{equation}
    \mathcal{L}_{\text{FM}}(\theta) = \mathbb{E}_{t,x_0,x_1} \| v_\theta(x_t,t) -(x_1-x_0) \|_2^2,
\end{equation}
where $x_1$ is the target video and $x_0$ is the Gaussian prior. The network $v_{\theta}$ learns to regress the vector field $x_1-x_0$ from the intermediate state $x_t=tx_1+(1-t)x_0$. This formulation corresponds to the flow ordinary differential equation:
\begin{equation}
    \frac{dx}{dt} = v_\theta(x,t).
\end{equation}

\subsection{SAMA}
SAMA is built upon the video diffusion model Wan2.1-T2V-14B~\cite{wan2025wan}. Given a source video $V_s$ and an editing instruction $y$, the goal is to generate an edited target video $V_t$ that follows $y$ while preserving realistic spatiotemporal motion and non-edited content.

\noindent\textbf{Latent tokenization.}
We encode videos into VAE latents following latent diffusion style formulations~\cite{rombach2021highresolution}. The source and target videos are represented as token sequences $\mathbf{z}_s$ and $\mathbf{z}_t$. We form an in-context V2V input by concatenating the source and (noisy) target token sequences: $\mathbf{z} = [\mathbf{z}_s \,;\, \mathbf{z}_t].$

\noindent\textbf{Type embeddings.} To disambiguate token roles, we add a learned type embedding to each token: type id $0$ for source-video latent tokens $\mathbf{z}_s$, type id $2$ for target-video latent tokens $\mathbf{z}_t$, and type id $1$ for semantic tokens introduced by Semantic Anchoring (Sec.~\ref{sec:sa}). This convention is used consistently across all stages. 
We empirically observe that using type embeddings leads to faster convergence than the commonly used shifted RoPE scheme~\cite{su2024roformer, song2025query}, while minimally perturbing the backbone prior. We provide further discussion and supporting evidence in the Appendix.

SAMA internalizes two complementary capabilities within the diffusion backbone: \emph{Semantic Anchoring (SA)} provides instruction-consistent anchors on sparse anchor frames to stabilize structural editing (see Sec.~\ref{sec:sa}); \emph{Motion Alignment (MA)} aligns the edited video with the source motion dynamics through motion-centric pretext supervision, improving temporal stability and mitigating semantic--motion conflicts (see Sec.~\ref{sec:ma}).
Building on these two capabilities, we further introduce a two-stage training strategy: we first learn strong inherent semantic--motion representations in a factorized pre-training stage, and then strengthen editing performance with paired supervision in an SFT stage (Sec.~\ref{sec:training}).

\subsection{Semantic Anchoring}
\label{sec:sa}
Semantic Anchoring (SA) is introduced as an auxiliary objective throughout both the \emph{Factorized Pre-training Stage} and the \emph{SFT Stage}.
For an image sample, the target image serves as the anchor.
For a video sample, we uniformly sample $N$ frames from the target video and treat them as sparse anchor frames.
Each anchor frame is encoded by a SigLIP image encoder~\cite{zhai2023sigmoid} to obtain patch-level semantic features.
We then aggregate these features into a compact token set by pooling, producing $M$ local semantic tokens that capture region-level semantics along with one global token that summarizes the overall content.
All semantic tokens are finally projected by a lightweight two-layer MLP into the same embedding space as the VAE latent tokens.

\emph{Injecting semantic tokens into the denoising sequence.} Let $\hat{\mathbf{s}}$ denote the projected semantic tokens extracted from the $N$ anchor frames.
We prepend $\hat{\mathbf{s}}$ to the target latent sequence and treat them as part of the denoising trajectory:
we apply the same forward noising process to both semantic tokens and target latents, and feed the concatenated noisy sequence into the DiT.
After denoising, we read out the positions corresponding to the semantic tokens and pass them through a semantic prediction head attached to the final DiT layer, yielding predicted semantic tokens $\mathbf{s}$.

\noindent\emph{Objective.}
We supervise semantic prediction with an $\ell_1$ loss between the predicted tokens and the extracted anchor tokens:
\begin{equation}
\mathcal{L}_{\text{sem}} = \|\hat{\mathbf{s}} - \mathbf{s}\|_1.
\end{equation}
The overall training objective combines the flow-matching loss and the Semantic Anchoring loss:
\begin{equation}
\mathcal{L} = \mathcal{L}_{\text{FM}} + \lambda \cdot \mathcal{L}_{\text{sem}}.
\label{equ:4}
\end{equation}

\begin{figure}[t]
  \centering
  \includegraphics[width=\linewidth]{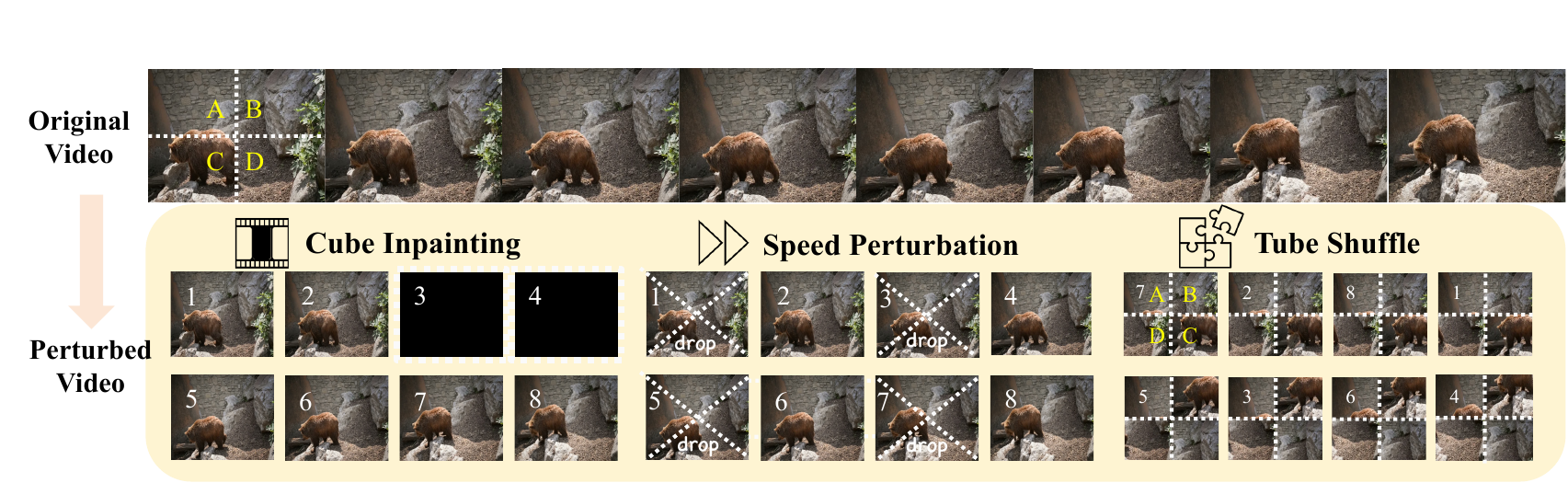} 
  \vspace{-6mm}
  \caption{\textbf{Illustration of pretext perturbations.} }
  \label{fig:2}
\end{figure}

\subsection{Motion Alignment}
\label{sec:ma}

Motion Alignment (MA) is applied on video samples in the \emph{Factorized Pre-training Stage} (Sec.~\ref{sec:training}). Given a source video $V_s$ and instruction $y$, we apply a motion-centric transformation $\mathcal{T}$ only to the source video to obtain $\tilde{V}_s=\mathcal{T}(V_s)$, while keeping the target side unchanged (i.e., always using the original target video without augmentation). This design forces the model to learn motion recovery and temporal reasoning from the source stream, improving robustness under fast motion and complex camera dynamics. Fig.~\ref{fig:2} provides an illustration of the pretext perturbations.

\noindent\emph{Motion-centric transformations.}
We adopt three restoration-style perturbations inspired by self-supervised learning for visual sequences~\cite{tong2022videomae, song2022ittakestwo, wang2023videomae2}: 
(i) \emph{Cube Inpainting}: mask a continuous temporal block in $\tilde{V}_s$ and recover missing content conditioned on the remaining frames; 
(ii) \emph{Speed Perturbation}: temporally accelerate $\tilde{V}_s$ and learn to restore normal dynamics, improving robustness to motion-rate changes; 
(iii) \emph{Tube Shuffle}: partition $\tilde{V}_s$ into a $2{\times}2{\times}2$ spatio-temporal tube grid and randomly permute tubes, forcing the model to reason about spatio-temporal structure and restore consistent motion. 

\noindent\emph{Prompting for pretext tasks.}
To make the objective explicit and unify the formulation across tasks, we prepend a short task token to the editing instruction:
\begin{center}
\begin{tcolorbox}[
  colback=gray!3,
  colframe=black!35,
  boxrule=0.4pt,
  arc=2pt,
  left=6pt,right=6pt,top=5pt,bottom=5pt,
  width=0.92\linewidth,
  title={Task token examples for Motion Alignment}
]
\small
\begin{itemize}\itemsep2pt
  \item \emph{[Complete the missing regions in the video.]} \quad (Cube Inpainting)
  \item \emph{[Restore the video to normal playback speed.]} \quad (Speed Perturbation)
  \item \emph{[Restore the correct spatio-temporal order of the video segments.]} \quad (Tube Shuffle)
\end{itemize}
\end{tcolorbox}
\end{center}

Overall, MA encourages the backbone to internalize robust motion dynamics from the source stream while remaining fully compatible with the instruction-conditioned editing formulation.

\subsection{Training Strategies}
\label{sec:training}
SAMA is optimized with a two-stage training pipeline that mirrors our factorized view of instruction-guided video editing.

\textbf{Stage~0: Factorized Pre-training.}
We start from a strong text-to-video prior and pre-train it on a mixture of instruction-based image editing pairs and large-scale text-to-video data~\cite{wang2025koala,hong2025motionbench}.
The image editing portion provides broad semantic coverage and improves general instruction grounding, while the text-to-video portion supplies diverse real-world motion patterns.
During this stage, we apply SA to both image and video samples, and apply MA only to the video stream:
(i) \emph{SA} supervises semantic token prediction on $N$ sparsely sampled anchor frames, encouraging instruction-consistent semantic anchoring while sharing the same diffusion backbone (Sec.~\ref{sec:sa});
(ii) \emph{MA} trains the model to restore temporally perturbed source videos with motion-centric pretext supervision, improving temporal stability and robustness under fast motion (Sec.~\ref{sec:ma}).
The overall objective at Stage~0 follows Eq.~(4),
\begin{equation}
\mathcal{L} = \mathcal{L}_{\text{FM}} + \lambda \cdot \mathcal{L}_{\text{sem}},
\end{equation}
where $\mathcal{L}_{\text{FM}}$ is the flow matching loss in Eq.~(1) and $\mathcal{L}_{\text{sem}}$ is the SA semantic prediction loss.

\textbf{Stage~1: Supervised Fine-tuning (SFT).}
We then perform supervised fine-tuning on paired video editing datasets~\cite{ditto,he2025openve,reco}, while mixing a small portion of image editing data to preserve general instruction-following behavior~\cite{nhredit,qian2025pico}.
In this stage, the model is trained on standard instruction-guided video editing triplets (source video, instruction, target video),
and we keep SA enabled to maintain stable semantic anchoring on sparse anchor frames.
Compared with Stage~0, Stage~1 focuses on aligning generation with paired editing supervision, improving edit fidelity and mitigating remaining semantic--motion conflicts observed in challenging motions and fine-grained edits.

This two-stage design separates the learning of semantic anchoring and motion alignment from scarce paired video-edit data. As a result, Stage~0 already provides strong zero-shot video editing capability, and Stage~1 further improves edit fidelity and benchmark performance with paired supervision.

%% file: sec/4_exp.tex
\section{Experiments}
\subsection{Experimental Settings}
\textbf{Training data.} 
As summarized in Tab.~\ref{tab:training_data_stages}, 
we use NHR-Edit~\cite{nhredit}, GPT-image-edit~\cite{gptimageedit}, X2Edit\cite{ma2025x2edit}, and Pico-Banana-400K~\cite{qian2025pico} for image editing training. 
We additionally incorporate text-to-video Koala-36M~\cite{wang2025koala} and MotionBench~\cite{hong2025motionbench} for pretext motion alignment.
Ditto-1M~\cite{ditto}, OpenVE-3M~\cite{he2025openve}, and ReCo-Data~\cite{reco} are employed for video editing. 
All datasets are additionally subjected to a VLM-based coarse filtering stage to remove low-quality or instruction-inconsistent samples. 
The detailed filtering criteria are provided in Appendix.
Specifically, we only use the Style subset of Ditto-1M~\cite{ditto}, and the Local Change, Background, Style, and Subtitles categories from OpenVE-3M~\cite{he2025openve}.

\input{table/dataset}

\textbf{Implementation details.} During training, we conduct two-stage training on mixed image and video data.
The learning rate is $2\times10^{-5}$ for both stages.
The global batch size is 448 for images and 112 for videos, and we train at a resolution of 480p. We support multiple aspect ratios, including $1/2,2/3,3/4,$ and $1/1$, as well as their reciprocals.
We maintain an exponential moving average (EMA~\cite{DDPM}) of model parameters with decay 0.9998 and update it every iteration.
The loss weight $\lambda$ (Eq.~\ref{equ:4}) is set to 0.1.
Unless otherwise specified, we uniformly sample $N$ sparse anchor frames for Semantic Anchoring (Sec.~\ref{sec:sa}); for efficiency, we set $N=1$ in all experiments. We use $M$ local semantic tokens per anchor frame (plus one global token), and fix $M=64$ throughout.

In the text-to-video data, we use no pretext task as well as three pretext tasks—Cube Inpainting, Speed Perturbation, and Tube Shuffle—with a sampling ratio of 1:2:3:4 (no-pretext : cube inpainting : speed perturbation : tube shuffle). 
Task-specific settings are deferred to Appendix.

\input{table/VIE-bench}

\input{table/OpenVE-bench}
\input{table/ReCo-bench}

\textbf{Evaluation details.} 
To evaluate SAMA, we compare it against current state-of-the-art methods, including closed-source and open-source systems. 
For closed-source models, we include Kling1.6~\cite{kling1.6}, Kling-Omni~\cite{klingomni}, Runway~\cite{runwayaleph}, MiniMax~\cite{zi2025minimax}, and Pika~\cite{pika}.
For open-source methods, we compare with InsV2V~\cite{insv2v}, DiffuEraser~\cite{diffueraser}, VACE~\cite{vace}, InsViE~\cite{insvie}, Omni-Video~\cite{tan2025omni}, LucyEdit~\cite{lucyedit}, UniVideo~\cite{univideo}, InstructX~\cite{instructx}, ICVE~\cite{icve}, Ditto~\cite{ditto}, OpenVE-Edit~\cite{he2025openve}, VINO~\cite{chen2026vino}, and ReCo~\cite{reco}. 
We conduct experiments on three benchmarks: VIE-Bench~\cite{instructx}, OpenVE-Bench~\cite{he2025openve}, and ReCo-Bench~\cite{reco}. 
We use different VLM judges for scoring across benchmarks: GPT-4o~\cite{gpt-4o} for VIE-Bench~\cite{instructx}, Gemini-2.5-Pro~\cite{comanici2025gemini2.5pro} for OpenVE-Bench~\cite{he2025openve}, and Gemini-2.5-Flash-Thinking~\cite{team2023gemini2.5flash} for ReCo-Bench~\cite{reco}.

\subsection{Comparisons with State-of-the-Art Methods}
Tab.~\ref{tab:VIE-bench} show that our method consistently outperforms existing open-source video-editing models across most metrics, while remaining competitive with state-of-the-art closed-source systems.
In particular, SAMA achieves the best overall performance on Swap/Change and Remove, among all compared methods.
Similar gains are observed on Tab. \ref{tab:openve_bench} on OpenVE-Bench~\cite{he2025openve} and Tab.\ref{tab:ReCo-bench} on ReCo-Bench~\cite{reco}, where SAMA attains the top overall score and delivers strong results across multiple task categories, despite a few metrics where it is not the best-performing method.

\begin{figure}[t!]
  \centering
  \includegraphics[width=0.99\textwidth]{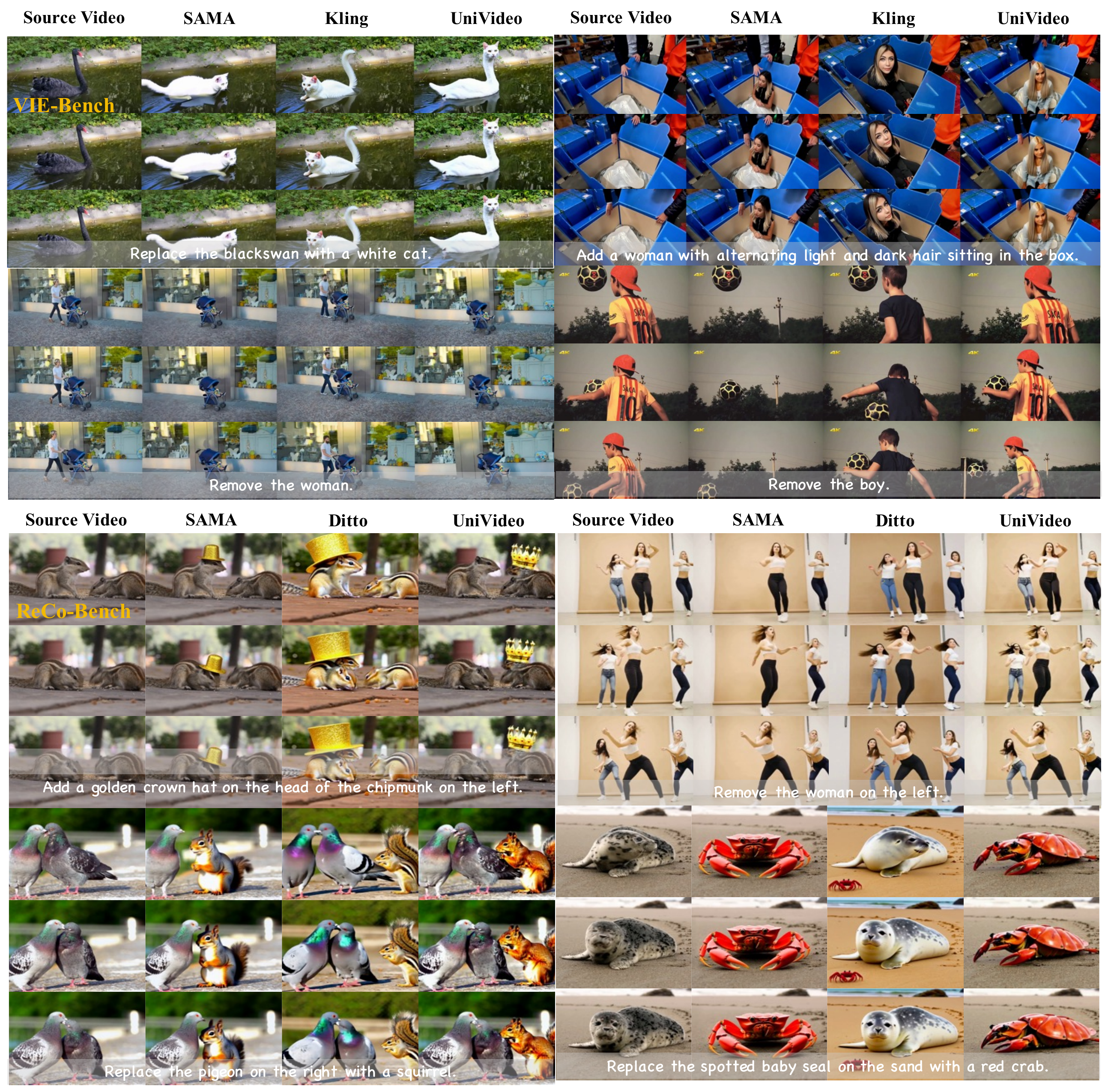} 
  \caption{Qualitative comparisons with prior methods on VIE-Bench and ReCo-Bench.}
  \label{fig:SOTA}
    \vspace{-0.2in}
\end{figure}

\textbf{Qualitative Comparisons.} In the qualitative comparisons on VIE-Bench and ReCo-Bench (see Fig.~\ref{fig:SOTA}), SAMA demonstrates stronger instruction adherence and temporal consistency across diverse editing types.
SAMA follows fine-grained instructions more reliably, correctly handling relative position cues (e.g., “on the left”) and attribute constraints (e.g., alternating \textit{light and dark hair}). It also completes replacements (e.g., pigeon$\rightarrow$squirrel, seal$\rightarrow$crab) with consistent appearance over time. Motion-wise, SAMA better preserves temporal alignment (e.g., keeping the stroller aligned after removal) and maintains identity/details during stylization, while other methods may drift or blur. 
Overall, SAMA better grounds instruction semantics while maintaining coherent motion, leading to higher-quality and more stable edits.
\emph{More qualitative results are provided in the Appendix.}

\begin{figure}[t]
  \centering
  \includegraphics[width=\linewidth]{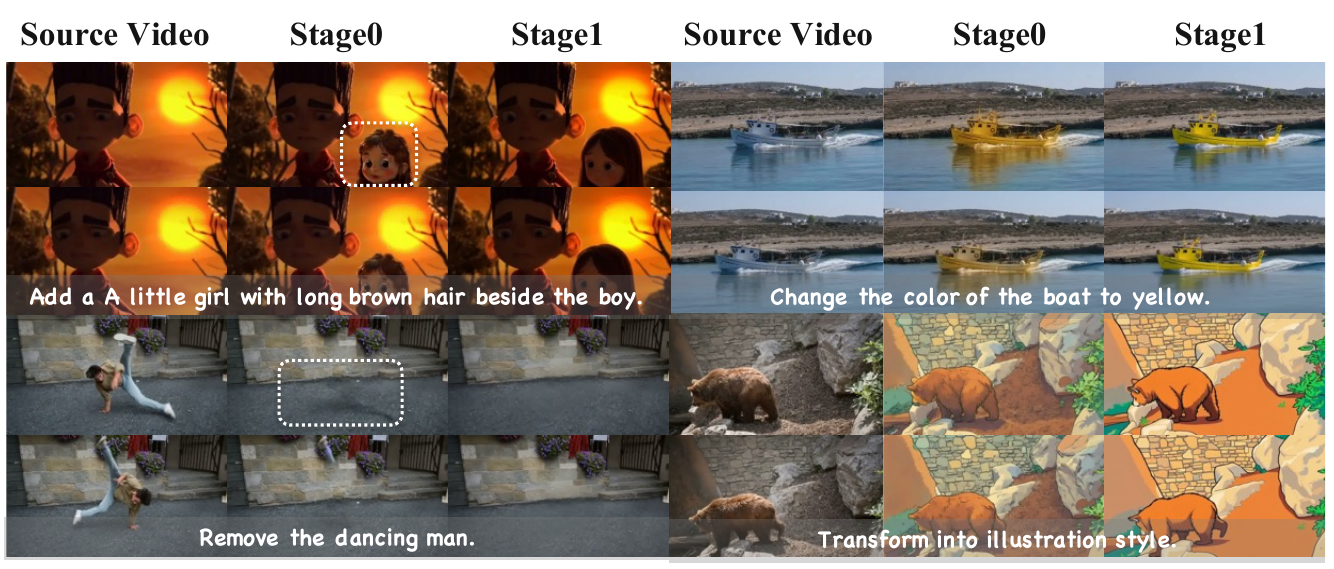}
  \caption{\textbf{Zero-shot qualitative results on VIE-Bench at two training stages.}}
  \label{fig:zero_shot}
\end{figure}


\subsection{Zero-shot Video Editing}
We evaluate SAMA in a zero-shot setting, where the model is trained w/o any video editing data and is directly prompted with editing instructions during inference. As show in Fig.~\ref{fig:zero_shot}, SAMA demonstrates strong zero-shot editing capabilities across Replace/Add/Remove/Style/Hybrid tasks, producing consistent edits over multiple frames while largely preserving non-edited content.
Despite these encouraging results, we also observe several typical failure modes in the zero-shot setting:
(i) attribute edits can be temporally inconsistent, e.g., the edited colors may vary across frames;
(ii) newly added objects may appear slightly blurry; 
(iii) removal edits may leave residual ghosting.

\subsection{Ablation Study}

\begin{figure}[t]
  \centering
  \begin{subfigure}[t]{0.46\linewidth}
    \centering
    \includegraphics[width=\linewidth]{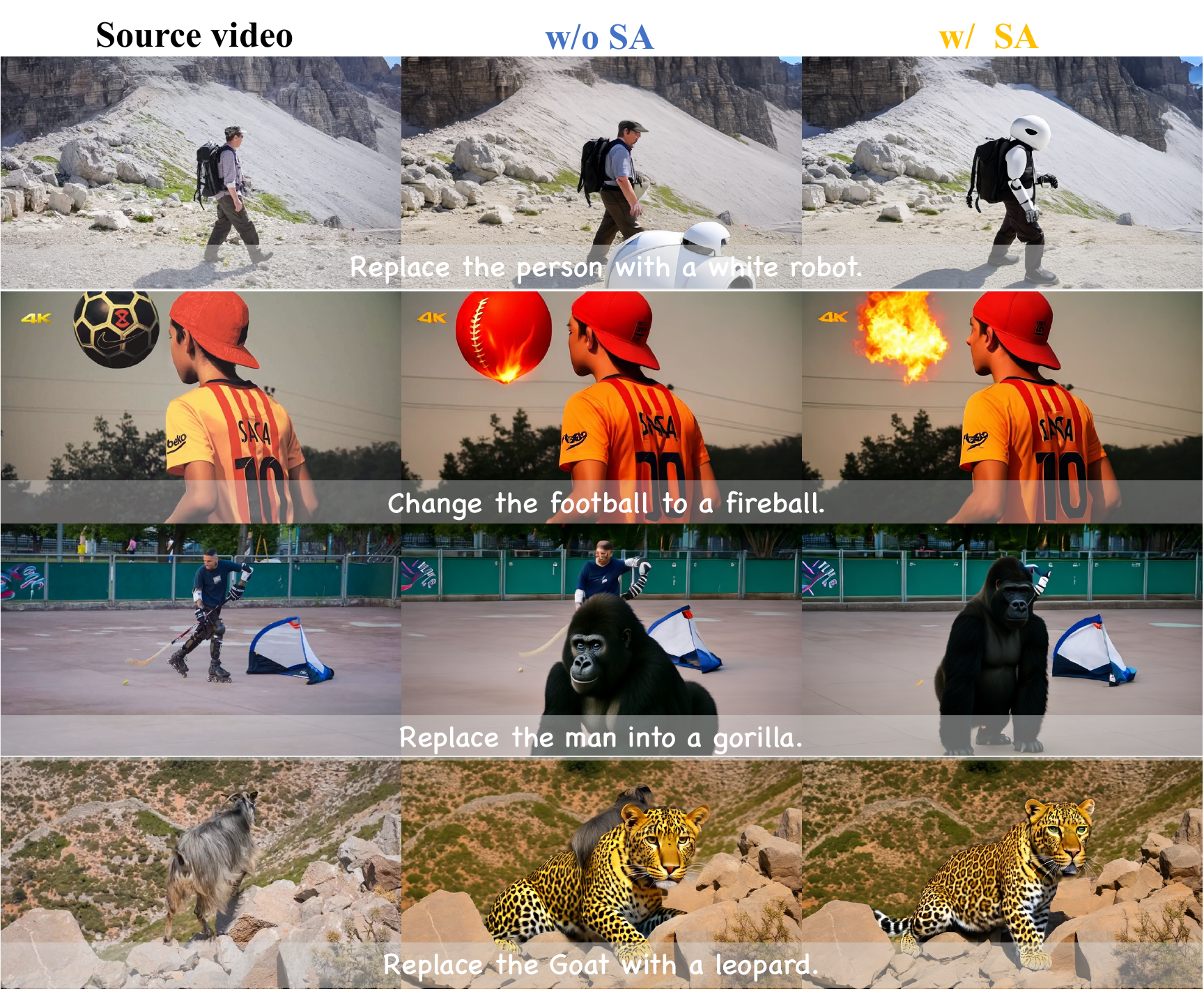}
    \caption{Visual results of SAMA w/ SA (right column) and w/o SA (middle column).
    }
    \label{fig:ablation_semantic_vis}
  \end{subfigure}
  \hfill
  \begin{subfigure}[t]{0.48\linewidth}
    \centering
    \includegraphics[width=\linewidth]{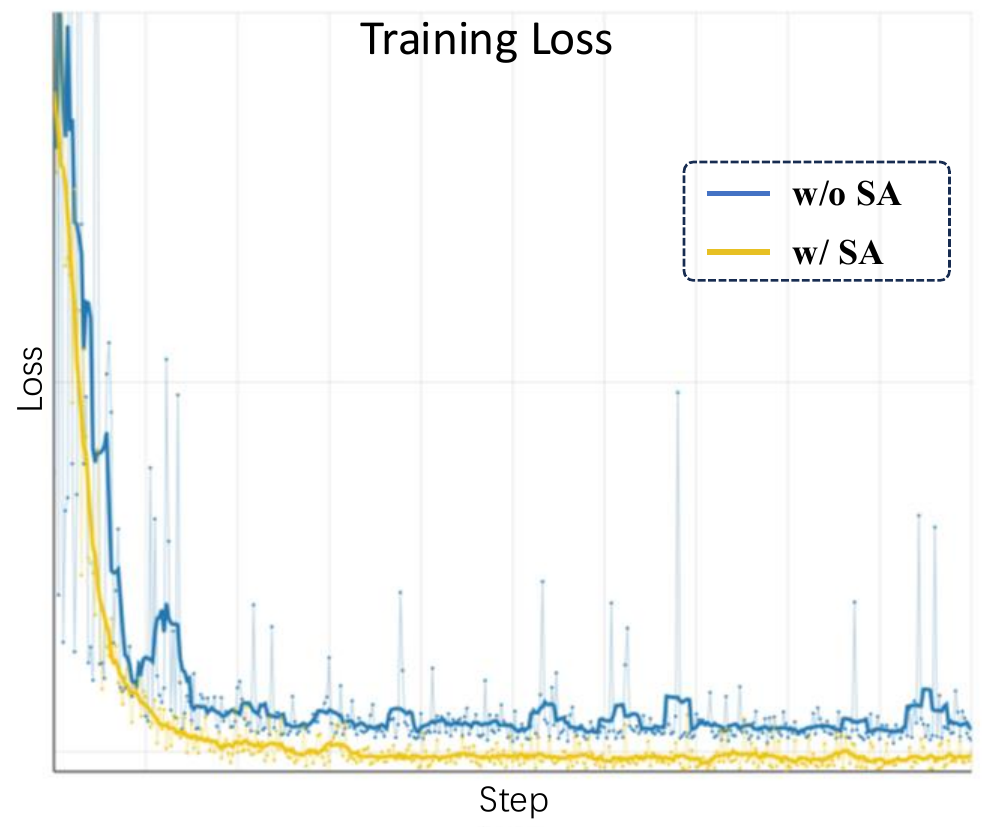}
    \caption{Training loss curves.}
    \label{fig:ablation_semantic}
  \end{subfigure}
  \caption{
  \textbf{Ablations for Semantic Anchoring (SA).}
  }
  \label{fig:ablation_semantic_pair}
\end{figure}

\begin{figure}[!b]
  \centering
  \includegraphics[width=\linewidth]{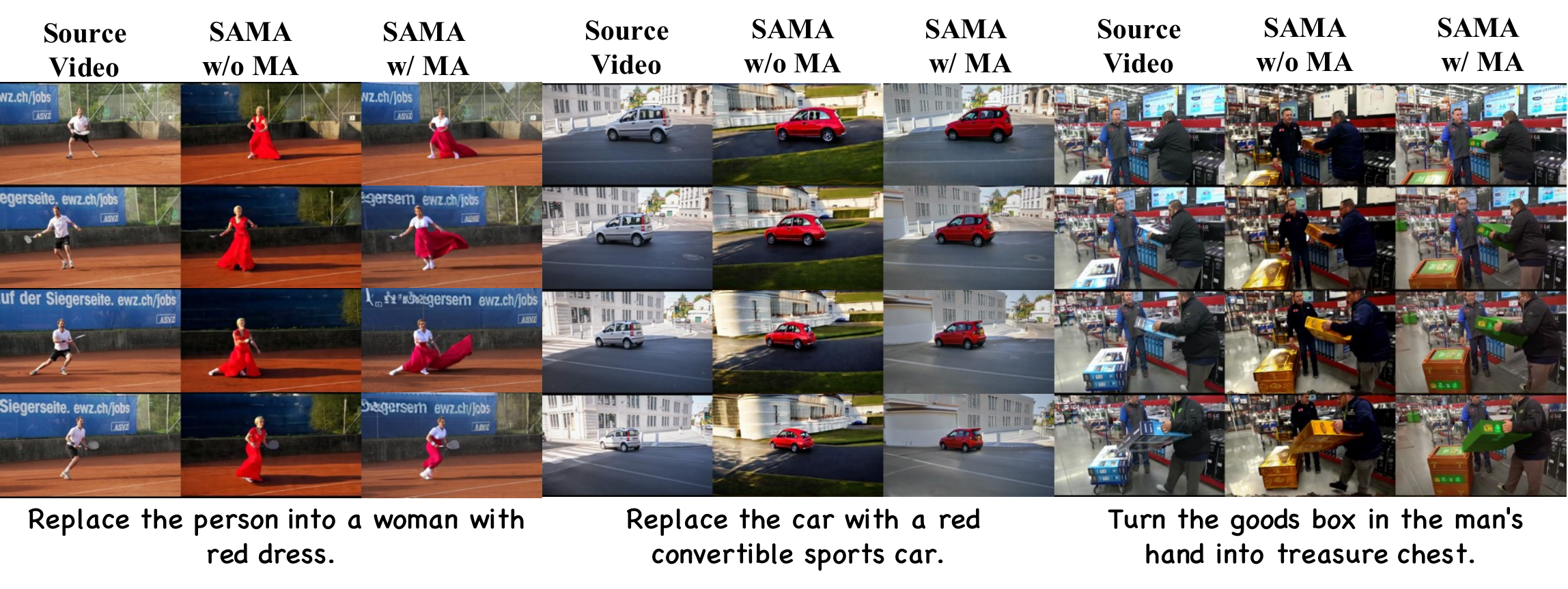}
  \caption{\textbf{Qualitative comparison of SAMA w/ and w/o \textit{MA}}.}
  \label{fig:ablation_pretext}
\end{figure}

\noindent\textbf{Semantic Anchoring}. We first observe that incorporating the semantic prediction objective accelerates the decrease of the diffusion loss, leading to faster DiT convergence. In addition, SA stabilizes training, as evidenced by a noticeably reduced loss variance (see Fig. ~\ref{fig:ablation_semantic}).
We set the baseline by concatenating the source latent with the video latent, without SA or MA. We use the smaller Wan2.2-T2V-5B~\cite{wanx} for efficiency with type embeddings and train it on a subset of the Ditto-1M~\cite{ditto}, obtaining the baseline results. Building on this baseline, adding SA leads to consistent mean score improvements across all tasks on VIE-Bench.

We further provide qualitative comparisons under the same number of training steps in Fig.~\ref{fig:ablation_semantic_vis}. As shown, the model with SA produces higher-quality edits at earlier training stages, whereas the baseline without it often yields incomplete or less accurate modifications. These results corroborate that SA facilitates faster convergence in practice.


\noindent\textbf{Motion Alignment}. We conduct a qualitative analysis on the effect of MA. We find that enabling MA improves temporal consistency under fast motion and alleviates motion blur. Representative qualitative results are shown in Fig.~\ref{fig:ablation_pretext}. 
\input{table/ablation}
In the tennis case with large camera motion, with MA noticeably improves background sharpness (e.g., clearer on-screen text), while the baseline appears blurred. Similar improvements are observed in the car and the third example, where the baseline often loses background motion. 
Quantitative ablation results on MA are summarized in Tab.~\ref{tab:ablation}. On VIE-Bench, adding MA alone improves the overall score by 0.399 over the baseline. When combining SA and MA, the overall score further increases by 0.783, indicating that the two components are complementary.



%% file: table/dataset.tex
\begin{table}[t]
\centering
\caption{
\textbf{Statistics of training data for each stage.}
$^{\bigstar}$ denotes we use specific text-to-video data for Motion Alignment by solving pretext transformations.
}
\label{tab:training_data_stages}
\setlength{\tabcolsep}{7pt} 
\resizebox{\linewidth}{!}{
\begin{tabular}{llrc}
\toprule
\textbf{Training stage} & \textbf{Dataset} & \textbf{\# Pairs} & \textbf{Type} \\
\midrule
\multirow{5}{*}{\makecell{\textbf{Stage 0} \\\textit{Factorized Pre-training}}}
& NHR-Edit~\cite{nhredit} & 720,087 & image editing \\
& GPT-Image-Edit~\cite{gptimageedit} & 1,015,170 & image editing \\
& X2Edit~\cite{ma2025x2edit} & 768,470 & image editing \\
& Koala-36M~\cite{wang2025koala} & 1,532,716 & text-to-video$^{\bigstar}$  \\
& MotionBench~\cite{hong2025motionbench} &53,879  & text-to-video$^{\bigstar}$ \\
\midrule
\multirow{5}{*}{\makecell{\textbf{Stage 1} \\\textit{Supervised Fine-tuning}}}
& NHR-Edit~\cite{nhredit} &  720,087  & image editing \\
& Pico-Banana-400K~\cite{qian2025pico} & 257,730 & image editing \\
& Ditto-1M~\cite{ditto} & 3,936 & video editing \\
& OpenVE-3M~\cite{he2025openve} & 818,232 & video editing \\
& ReCo-Data~\cite{reco} & 206,596 & video editing \\
\bottomrule
\end{tabular}
}

\end{table}

%% file: table/VIE-bench.tex
\begin{table}[t]
\centering
\caption{\textbf{Comparison results on VIE-Bench}. The best results are shown in \textbf{bold}. Gray shading indicates closed-source models.}
\label{tab:VIE-bench}
\footnotesize
\resizebox{\textwidth}{!}{
\begin{tabular}{l| c c c c | c c c c}
\toprule
\multirow{2}{*}{\textbf{Method}}
& \makecell{Instruct\\follow} & \makecell{Preser-\\vation} & \makecell{Quality} & \makecell{Avg.}
& \makecell{Instruct\\follow} & \makecell{Preser-\\vation} & \makecell{Quality} & \makecell{Avg.} \\
\cline{2-9}
& \multicolumn{4}{c|}{\textbf{Add}} & \multicolumn{4}{c}{\textbf{Swap / Change}} \\
\midrule
\rowcolor{BoxBackground} Kling1.6 & 6.000 & 8.230 & 5.576 & 6.602 & 9.000 & 9.060 & 8.333 & 8.800 \\
\rowcolor{BoxBackground} Kling-Omni & 9.333 & 9.589 & 8.622 & 9.181 & 9.495 & 9.448 & 8.638 & 9.194 \\
\rowcolor{BoxBackground} Runway & 8.607 & 8.913 & 7.823 & 8.447 & 9.580 & 8.628 & 9.275 & 9.161 \\
\rowcolor{BoxBackground} Pika & - & - & - & - & 7.542 & 7.847 & 6.837 & 7.408 \\
InsV2V & 3.552 & 5.891 & 3.402 & 4.281 & 5.304 & 6.428 & 4.971 & 5.567 \\
VACE & 3.938 & 6.696 & 3.929 & 4.854 & 6.171 & 7.552 & 6.199 & 6.640 \\
Omni-Video & 5.699 & 6.135 & 6.294 & 6.242 & 4.733 & 4.856 & 4.656 & 4.748 \\
UniVideo & \textbf{8.567} & 9.422 & 7.978 & 8.656 & 8.886 & 8.962 & 8.200 & 8.683 \\
InstructX & 8.446 & 8.683 & 7.919 & 8.349 & 9.514 & 9.171 & 8.533 & 9.072 \\
\rowcolor{oursrow} \textbf{SAMA} & 8.467 & \textbf{9.422} & \textbf{8.244} & \textbf{8.711} & \textbf{9.733} & \textbf{9.514} & \textbf{8.771} & \textbf{9.340} \\
\midrule
& \multicolumn{4}{c|}{\textbf{Remove}} & \multicolumn{4}{c}{\textbf{Style / Tone Change}} \\
\midrule
\rowcolor{BoxBackground} Kling1.6 & 8.440 & 8.800 & 7.520 & 8.253 & - & - & - & - \\
\rowcolor{BoxBackground} Kling-Omni & 9.378 & 9.233 & 8.789 & 9.133 & 9.867 & 9.200 & 8.956 & 9.341 \\
\rowcolor{BoxBackground} Runway & 8.664 & 9.145 & 7.703 & 8.504 & 9.583 & 9.200 & 8.616 & 9.133 \\
MiniMax & 6.963 & 7.518 & 6.037 & 6.839 & - & - & - & - \\
DiffuEraser & 6.346 & 6.807 & 5.576 & 6.243 & - & - & - & - \\
InsV2V & 1.209 & 3.769 & 1.322 & 2.098 & 7.835 & 8.086 & 6.437 & 7.452 \\
VACE & 1.812 & 3.877 & 2.359 & 2.682 & - & - & - & - \\
Omni-Video & 6.004 & 5.970 & 4.807 & 5.593 & 5.486 & 4.655 & 5.959 & 5.366 \\
UniVideo & 8.133 & 8.778 & 7.789 & 8.233 & 9.244 & 8.689 & 8.200 & 8.711 \\
InstructX & 8.627 & 8.668 & 7.672 & 8.322 & \textbf{9.650} & 9.099 & \textbf{8.839} & 9.196 \\
\rowcolor{oursrow} \textbf{SAMA} & \textbf{9.533} & \textbf{9.189} & \textbf{8.711} & \textbf{9.144} & 9.644 & \textbf{9.356} & 8.778 & \textbf{9.259} \\
\bottomrule
\end{tabular}
}
\end{table}

%% file: table/OpenVE-bench.tex
\begin{table}[t]
\centering
\caption{\textbf{Comparison results on OpenVE-Bench with Gemini 2.5 Pro}. The best results are highlighted in \textbf{bold}. Gray shading indicates closed-source models.
}
\label{tab:openve_bench}
\setlength{\tabcolsep}{3pt} 
\renewcommand{\arraystretch}{1.05} 

\begin{adjustbox}{max width=0.95\textwidth}
\begin{tabular}{lc|ccccccc}
\toprule
\textbf{Method} & \textbf{\# Params.}   & \textbf{\makecell{Global\\ Style} }& \textbf{\makecell{Background \\Change}} & \textbf{\makecell{Local\\ Change}} & \textbf{\makecell{Local\\ Remove} }& \textbf{\makecell{Local\\ Add} }& \textbf{\makecell{Subtitle\\ Edit} }& \textbf{\makecell{Creative\\ Edit} }\\
\midrule
\rowcolor{BoxBackground}Runway      & -      & 3.72 & 2.62 & 4.18 & 4.16 & 2.78 & 3.62 & 3.64  \\
VACE              & 14B    & 1.49 & 1.55 & 2.07 & 1.46 & 1.26 & 1.48 & 1.47  \\
Omni-Video         & 1.3B   & 1.11 & 1.18 & 1.14 & 1.14 & 1.36 & 1.00 & 2.26  \\
InsViE            & 2B     & 2.20 & 1.06 & 1.48 & 1.36 & 1.17 & 2.18 & 2.02  \\
Lucy-Edit         & 5B     & 2.27 & 1.57 & 3.20 & 1.75 & 2.30 & 1.61 & 2.86  \\
ICVE              & 13B   & 2.22 & 1.62 & 2.57 & 2.51 & 1.97 & 2.09 & 2.41  \\
Ditto             & 14B    & 4.01 & 1.68 & 2.03 & 1.53 & 1.41 & 2.81 & 1.23  \\
OpenVE-Edit       & 5B    & 3.16 & 2.36 & 2.98 & 1.85 & 2.15 & 2.91 & 2.31  \\
UniVideo       &20B      & 3.64 & 2.22 & 3.91 & 2.70 & \textbf{2.98} & 2.69 & 2.90  \\
\rowcolor{oursrow}
\textbf{SAMA}             & 14B    &\textbf{4.05}  &\textbf{2.59}  &\textbf{3.93}    &\textbf{3.32}  &2.54  &\textbf{3.63}  &\textbf{3.11}    \\
\bottomrule
\end{tabular}
\end{adjustbox}
\end{table}

%% file: table/ReCo-bench.tex
\begin{table}[t]
	\centering
	\caption{\textbf{Comparison results on ReCo-Bench with Gemini-2.5-Flash-Thinking}. The best results are shown in \textbf{bold}. Abbreviations: SA, semantic accuracy; SP, scope precision; CP, content preservation; AN, appearance naturalness; SN, scale naturalness; MN, motion naturalness; VF, visual fidelity; TS, temporal stability; ES, edit stability. $S_{EA}$/$S_{VN}$/$S_{VQ}$ are category scores and $S$ is the overall score. 
		}
	\label{tab:ReCo-bench}
	\setlength{\tabcolsep}{2pt}
	\resizebox{\textwidth}{!}{%
		\begin{tabular}{@{}c c | ccc | ccc | ccc | cccc@{}}
			\toprule
			\multirow{2}{*}{\textbf{Task}} & \multirow{2}{*}{\textbf{Method}} &
                \multicolumn{3}{c|}{\makecell{\textbf{E}dit \textbf{A}ccuracy}} &
			\multicolumn{3}{c|}{\makecell{\textbf{V}ideo \textbf{N}aturalness}} &
			\multicolumn{3}{c|}{\makecell{\textbf{V}ideo \textbf{Q}uality}} &
			\multicolumn{4}{c}{\makecell{\textbf{Average Score}}} \\
			\cmidrule{3-15}
			& &
			{SA} & {SP} & {CP} &
			{AN} & {SN} & {MN} &
			{VF} & {TS} & {ES} &
			$S_{EA}$ & $S_{VN}$ & $S_{VQ}$ & $S$ \\

			\midrule
			\multirow{7}{*}{\textbf{Add}} 
			& InsViE & 2.60 & 2.79 & 2.78 & 2.33 & 3.98 & 3.74 & 3.71 & 3.91 & 3.58 & 2.60 & 3.10 & 3.46 & 3.05  \\
			& Lucy-Edit & 6.27 & 6.32 & 7.75 & 4.63 & 7.08 & 6.08 & 6.31 & 6.82 & 7.57 & 6.47 & 5.70 & 6.77 & 6.31 \\
			& Ditto & 7.46 & 7.24 & 6.30 & 6.30 & 8.85 & 8.30 & 8.13 & 8.55 & 9.03 & 6.70 & 7.57 & 8.41 & 7.56 \\
                &  UniVideo & 9.39&\textbf{9.27}&9.69&7.27&9.23 &8.80 & 8.44 & 8.89 & 9.75 &9.40 &8.31 &8.99 &8.90 \\
                &  ReCo & 8.65&8.40& 9.22 & 6.39 & 8.78 & 8.28 &  8.02 &  8.61 & 9.61 & 8.54 & 7.55 & 8.61 & 8.23 \\
                & \cellcolor{oursrow} \textbf{SAMA} & 
                \cellcolor{oursrow} \textbf{9.51} & \cellcolor{oursrow}9.26 & \cellcolor{oursrow} \textbf{9.83} & \cellcolor{oursrow} \textbf{7.44} & \cellcolor{oursrow} \textbf{9.50} & \cellcolor{oursrow} \textbf{8.87} & \cellcolor{oursrow} \textbf{8.78} & \cellcolor{oursrow} \textbf{9.03} & \cellcolor{oursrow}\textbf{9.76} & \cellcolor{oursrow} \textbf{9.43} & \cellcolor{oursrow} \textbf{8.33} & \cellcolor{oursrow} \textbf{9.00} & \cellcolor{oursrow} \textbf{8.92} \\
			\midrule
            
			\multirow{8}{*}{\textbf{Replace}} 
			& InsViE & 1.89 & 2.38 & 2.48 & 2.58 & 5.25 & 5.05 & 3.76 & 4.00 & 3.52 & 2.10 & 3.91 & 3.49 & 3.17  \\
			& Lucy-Edit & 6.57 & 7.49 & 7.73 & 5.13 & 7.46 & 6.65 & 6.32 & 6.64 & 8.08 & 7.08 & 6.21 & 6.88 & 6.72 \\
			& Ditto & 4.95 & 4.83 & 4.79 & 5.81 & 8.63 & 8.10 & 7.55 & 7.95 & 8.71 & 4.56 & 7.21 & 7.96 & 6.58  \\
                &  UniVideo &  9.03 & 9.68 &  9.73& 7.73 &9.30 & 8.92 &\textbf{8.57} &\textbf{8.91} &\textbf{9.80} &  9.40 & 8.39  &8.90&8.90  \\
                &  ReCo & 9.38 &  9.43 &  9.59 &  7.07 &  8.87 &  8.47 &  8.19 &  8.65 &  9.67 &9.43 &  8.01 &  8.77 &   8.74 \\
                 & \cellcolor{oursrow} \textbf{SAMA} & \cellcolor{oursrow} \textbf{9.58} & \cellcolor{oursrow} \textbf{9.82} & \cellcolor{oursrow} \textbf{9.82} & \cellcolor{oursrow} \textbf{7.77} & \cellcolor{oursrow} \textbf{9.35} & \cellcolor{oursrow} \textbf{8.98} & \cellcolor{oursrow}8.55 & \cellcolor{oursrow}8.80 & \cellcolor{oursrow} 9.72 &
			\cellcolor{oursrow} \textbf{9.71} & \cellcolor{oursrow} \textbf{8.60} & \cellcolor{oursrow} \textbf{8.98} &  \cellcolor{oursrow} \textbf{9.10} \\
			\midrule
            
			\multirow{6}{*}{\textbf{Remove}}  
			& InsViE & 2.53 & 2.49 & 2.44 & 2.63 & 4.87 & 4.72 & 3.41 & 3.67 & 3.40 & 2.44 & 3.76 & 3.29 & 3.16 \\
			& VACE& 4.58 & 4.58 & 4.56 & 4.96 & 6.09 & 5.89 & 5.48 & 5.50 & 5.57 & 4.57 & 5.43 & 5.56 & 5.19  \\
                & UniVideo & 7.37 & 7.43 & 7.28 & 6.06 & 7.61 & 7.13 & 6.28 & 6.43 & 7.72 & 7.33 & 6.59 &  6.51 & 6.81 \\
                & ReCo & 7.43 & 7.43 & 7.17 & 6.20 & 7.43 & 7.30 & 6.48 & 6.63 & 7.68 & 7.28 & 6.90 &  6.82 & 7.00 \\
                & \cellcolor{oursrow} \textbf{SAMA} & \cellcolor{oursrow} \textbf{8.76} & \cellcolor{oursrow} \textbf{8.71} & \cellcolor{oursrow} \textbf{8.43} & \cellcolor{oursrow} \textbf{7.16} & \cellcolor{oursrow} \textbf{8.73} & \cellcolor{oursrow} \textbf{8.42} & \cellcolor{oursrow} \textbf{7.31} & \cellcolor{oursrow} \textbf{7.52} & \cellcolor{oursrow} \textbf{8.73} & \cellcolor{oursrow} \textbf{8.61}  &  \cellcolor{oursrow} \textbf{7.94} & \cellcolor{oursrow} \textbf{7.73} & \cellcolor{oursrow} \textbf{8.09}\\
			\midrule
            
			\multirow{6}{*}{\textbf{Style}}  
                & InsViE & 7.59 & 8.86 & 8.49 & 6.77 & 9.14 & 9.28 & 7.13 & 6.40 & 8.99 & 8.17 & 8.21 & 7.35 & 7.91 \\
			& Lucy-Edit & 3.73 & 5.59 & 5.39 & 4.20 & 5.88 & 5.88 & 4.44 & 4.17 & 5.87 & 4.65 & 4.67 & 5.17 & 4.83 \\
			& Ditto &9.10 & 9.36 & 9.26 & 8.25 & 9.51 & 9.58 & 8.33 & 8.33 & 9.77 & 9.20 & 9.07 & 8.77 & 9.01  \\
                & UniVideo &  8.10 &  9.82&  9.50&  8.56& 9.65&  \textbf{9.84}&  \textbf{8.91}& 8.57&  \textbf{9.88}&  8.95& 9.23&  9.00&  9.06\\
                &  ReCo &  \textbf{9.11} &  9.82 &  9.54 &  8.43 &  9.55 &  9.70 &  8.61 &  8.35 & 9.87 &  \textbf{9.42} &  9.19 &  8.90 &  9.17\\
                & \cellcolor{oursrow} \textbf{SAMA} & \cellcolor{oursrow} 8.46 & \cellcolor{oursrow} \textbf{9.95} & \cellcolor{oursrow} \textbf{9.64} & \cellcolor{oursrow} \textbf{8.79} & \cellcolor{oursrow} \textbf{9.77} & \cellcolor{oursrow} 9.77 & \cellcolor{oursrow} 8.88 & \cellcolor{oursrow} \textbf{8.59} & \cellcolor{oursrow} 9.83 & \cellcolor{oursrow}9.24 & \cellcolor{oursrow} \textbf{9.42} & \cellcolor{oursrow} \textbf{9.07} & \cellcolor{oursrow} \textbf{9.25} \\
			\bottomrule
		\end{tabular}
	}
\end{table}

%% file: table/ablation.tex

\begin{wraptable}{r}{0.57\linewidth}
\vspace{-22pt} 
\centering
\small
\caption{\textbf{Ablations of SAMA modules.}}
\vspace{-3pt}
\label{tab:ablation}
\setlength{\tabcolsep}{4pt}      
\renewcommand{\arraystretch}{1.05} 
\resizebox{0.9\linewidth}{!}{
\begin{tabular}{@{}lcccc@{}}
\toprule
method & \makecell{instruct\\follow} & \makecell{preser-\\vation} & quality & \makecell{Overall} \\
\midrule
baseline & 6.575 & 6.261 & 6.100 & 6.312 \\
w/ SA & 7.002 & 6.744 & 6.342 & 6.696 \\
w/ MA & 6.969 & 6.620 & 6.544 & 6.711 \\
\textbf{SAMA} & \textbf{7.402} & \textbf{6.998} & \textbf{6.884} & \textbf{7.095} \\
\bottomrule
\end{tabular}
}
\vspace{-16pt} 
\end{wraptable}

%% file: sec/5_conclusion.tex
\section{Conclusion}
We presented \textbf{SAMA}, a factorized framework for instruction-guided video editing that separates semantic anchoring and motion alignment within a DiT. 
Semantic anchoring introduces an explict prior via semantic-token prediction at anchor frames, while motion alignment improves temporal coherence through motion-centric restoration pre-training on text-to-video data. 
Extensive experiments on VIE-Bench, OpenVE-Bench, and ReCo-Bench demonstrate state-of-the-art performance among open-source methods and competitive results against commercial systems.
Moreover, SAMA exhibits strong zero-shot editing behavior, suggesting that robust instruction following can emerge from learning disentangled semantic and motion representations.
Future work will focus on long-video editing, fast-motion scenarios, and stronger semantic tokenization to further reduce residual artifacts and temporal inconsistencies.

%% file: sec/7_appendix.tex
\section{Discussion on Type Embeddings vs. Shifted RoPE}
\label{sec:rope}
\begin{figure}[htbp]
  \centering
  \includegraphics[width=\linewidth]{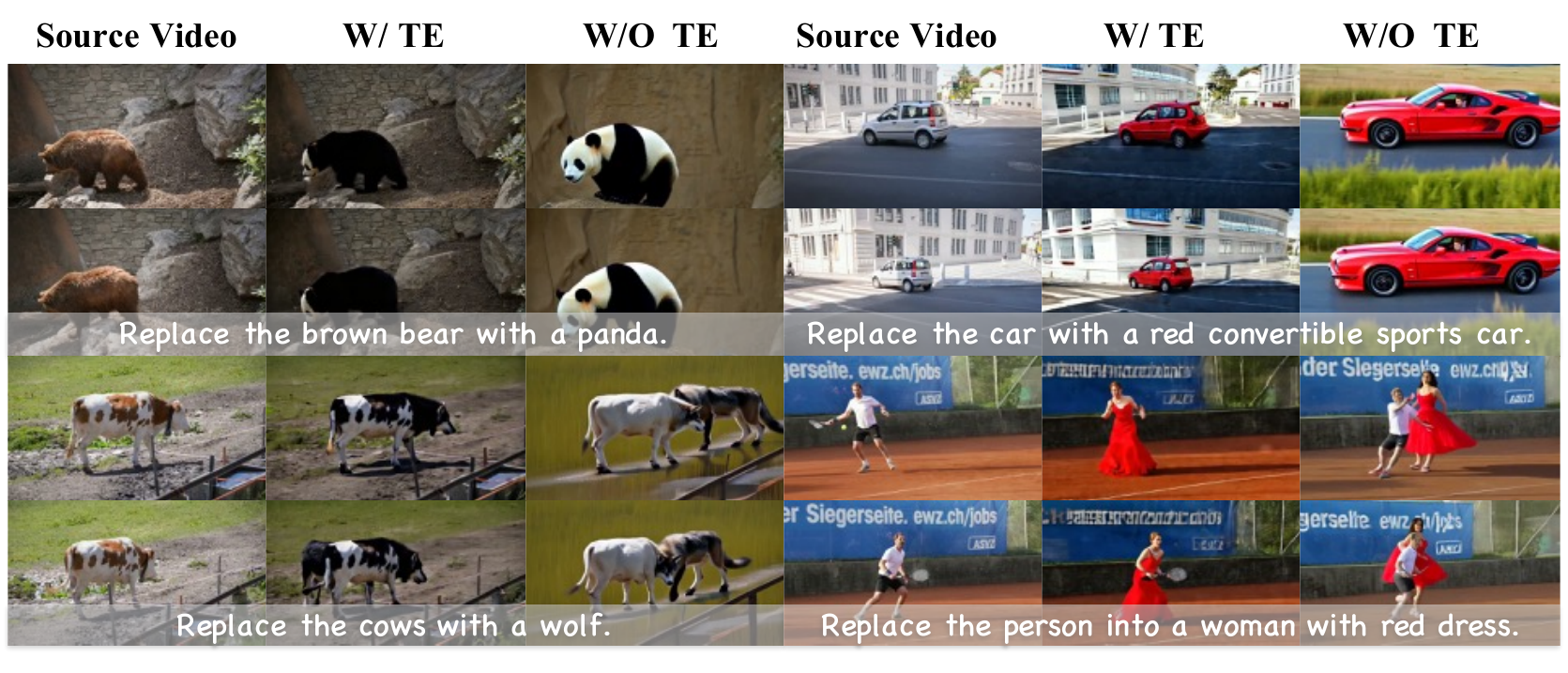} 
  \caption{\textbf{Illustration of Type Embedding (TE).} }
  \label{fig:1}
\end{figure}

To distinguish token roles in our unified formulation, we add a learned type embedding to each token (source-video latents, target-video latents, and semantic tokens), and apply this design throughout all training stages. We adopt type embeddings because they provide an explicit yet lightweight way to encode token identity without altering the backbone’s positional encoding, and introduce a smaller perturbation to the pretrained prior than shifted RoPE. Empirically, type embeddings yield faster and more stable convergence than shifted RoPE, likely because they decouple token role from token position: positional encoding continues to capture spatiotemporal structure, while token identity is modeled separately. Additional evidence is provided in Fig.~\ref{fig:1} and Tab.~\ref{tab:ablation}. Under the Wan2.2-T2V-5B~\cite{wanx} LoRA setting, training on the Ditto-1M replace subset~\cite{ditto} and evaluating on VIE-Bench replace~\cite{instructx}, type embeddings converge faster and better preserve background content.

\begin{wraptable}{r}{0.59\linewidth}
\vspace{-10pt} 
\centering
\small
\caption{\textbf{Ablations of Type Embeddings (TE) modules.}}
\label{tab:ablation}
\setlength{\tabcolsep}{4pt}      
\renewcommand{\arraystretch}{1.05} 
\resizebox{0.9\linewidth}{!}{
\begin{tabular}{@{}lcccc@{}}
\toprule
method & \makecell{instruct\\follow} & \makecell{preser-\\vation} & quality & \makecell{Overall} \\
\midrule
w/ PE & 6.705 & 7.533 & 6.686 & 6.975 \\
w/o PE & 6.619 & 6.257 & 6.619 & 6.498 \\
\bottomrule
\end{tabular}
}
\vspace{-4pt} 
\end{wraptable}

\section{VLM-based Data Filtering Details}
\label{sec:filter}
This appendix provides additional details on the training data used in our pipeline.
For data filtering, we use Qwen2.5-VL-72B~\cite{bai2025qwen25vltechnicalreport} as a VLM judge to score each sample from 1-10 via three inference turns, and average the scores. Scores are on a 1–10 scale. The detailed prompts are as follows. 
The judge assigns four scores: Instruction Following, Visual Quality, Content Preservation, and Motion Consistency (only for videos).
We then filter samples with the following thresholds.
For images, we use samples with Instruction Following $\ge 9$, Visual Quality $\ge 9$, and Content Preservation $\ge9$.
For videos, we use samples with Instruction Following $\ge 8$, Visual Quality $\ge 9$, Content Preservation $\ge 8$, and Motion Consistency $\ge 8$.

\begin{tcolorbox}[
    breakable,
    enhanced,
    width=\textwidth,
    title={Prompt for VLM-based Video Editing Evaluation: Instruction Following and Visual Quality},
    colback=mygray,
    colframe=black,
    boxrule=0.6pt,
    sharp corners,
    left=0mm,
    right=0mm,
    top=1mm,
    bottom=1mm,
    boxsep=1mm
]
\begin{lstlisting}[basicstyle=\ttfamily\footnotesize,breaklines=true,breakatwhitespace=false,columns=fullflexible,xleftmargin=0pt,xrightmargin=0pt,aboveskip=0pt,belowskip=0pt]
You are an expert video editing evaluator. Please evaluate the editing quality.

# User's Editing Instruction:
{instruction}

# Videos to Compare:
1. First video: Original video (before editing)
2. Second video: Edited video (after editing)

# Evaluation Criteria (Rate 1-10):
1. Instruction Following: How well does the edited video implement the user's specific instruction?
- 10: Perfectly follows all aspects
- 7-9: Mostly follows with minor issues
- 4-6: Partially follows, some aspects missing
- 1-3: Does not follow the instruction

2. Visual Quality: Is the edited video visually coherent and natural-looking?
- 10: Perfect quality, seamless editing
- 7-9: High quality, minor artifacts
- 4-6: Moderate quality, noticeable issues
- 1-3: Poor quality, obvious problems

# Output Format:
Return ONLY a JSON object with this exact structure:
{
  "instruction_following_score": <number between 1 and 10>,
  "visual_quality_score": <number between 1 and 10>,
  "explanation": "<brief explanation for the scores>"
}

Important: Only output the JSON object, nothing else.
\end{lstlisting}
\end{tcolorbox}

\begin{tcolorbox}[
    breakable,
    enhanced,
    width=\textwidth,
    title={Prompt for VLM-based Video Editing Evaluation: Motion Consistency},
    colback=mygray,
    colframe=black,
    boxrule=0.6pt,
    sharp corners,
    left=0mm,
    right=0mm,
    top=1mm,
    bottom=1mm,
    boxsep=1mm
]
\begin{lstlisting}[basicstyle=\ttfamily\footnotesize,breaklines=true,breakatwhitespace=false,columns=fullflexible,xleftmargin=0pt,xrightmargin=0pt,aboveskip=0pt,belowskip=0pt]
You are an expert video editing evaluator. Please evaluate the editing quality.
# User's Editing Instruction:
{instruction}
# Videos to Compare:
1. First video: Edited video (after editing)
2. Second video: Original video (before editing)
# Evaluation Criteria (Rate 1-10):
1. **Motion Consistency**: Does the edited video preserve the original motion of both the main object and the background?
Note: Any style, object, appearance changes are acceptable. You should only care about motion inconsistency which NOT caused by style, object, appearance change.
- 10: Perfectly preserves all original motions
- 7-9: Mostly preserves, minor motion deviations
- 4-6: Noticeable motion inconsistencies, or noticeable motion delays
- 4-6: Unnatural texture in the background of the edited video
- 1-3: Most of motion mismatches
- 1-3: In edited video, the first frame differs significantly from subsequent frames

# Output Format:
Return ONLY a JSON object with this exact structure:
{{
    "motion_consistency_score": <number between 1 and 10>,
    "explanation": "<brief explanation for the scores>"
}}

Important: Only output the JSON object, nothing else.
\end{lstlisting}
\end{tcolorbox}

\begin{tcolorbox}[
    breakable,
    enhanced,
    width=\textwidth,
    title={Prompt for VLM-based Video Editing Evaluation: Content Preservation},
    colback=mygray,
    colframe=black,
    boxrule=0.6pt,
    sharp corners,
    left=0mm,
    right=0mm,
    top=1mm,
    bottom=1mm,
    boxsep=1mm
]
\begin{lstlisting}[basicstyle=\ttfamily\footnotesize,breaklines=true,breakatwhitespace=false,columns=fullflexible,xleftmargin=0pt,xrightmargin=0pt,aboveskip=0pt,belowskip=0pt]
You are an expert video editing evaluator. Please evaluate the editing quality. Specifically, please evaluate the Content Preservation of the edited video only.

# User's Editing Instruction:
{instruction}

# Videos to Compare:
1. First video: Original video (before editing)
2. Second video: Edited video (after editing)

# Evaluation Criteria (Rate 1-10):
1. **Content Preservation**: Does the edit preserve all unrelated content from the original video?
    - 10: Perfectly preserves all unrelated content
    - 7-9: Mostly preserves, minor changes
    - 4-6: Some unrelated changes
    - 1-3: Changes much unrelated content
    
# Output Format:
Return ONLY a JSON object with this exact structure:
{{
    "content_preservation_score": <number between 1 and 10>,
    "explanation": "<brief explanation for the scores>"
}}

Important: Only output the JSON object, nothing else.
\end{lstlisting}
\end{tcolorbox}

\section{Task-Specific Settings for Text-to-Video Pretext Tasks}
As mentioned in the main text, for text-to-video data, we use no pretext task together with three pretext tasks: Cube Inpainting, Speed Perturbation, and Tube Shuffle. 

For Cube Inpainting, we use a masking ratio of 30\%. 
For Speed Perturbation, we apply a $2\times$ temporal acceleration. 
For Tube Shuffle, we divide each video into $2\times2\times2$ spatiotemporal tubes and randomly shuffle them.

\section{Pretext Prediction Visualization}
\begin{figure}[t]
  \centering
  \includegraphics[width=\linewidth]{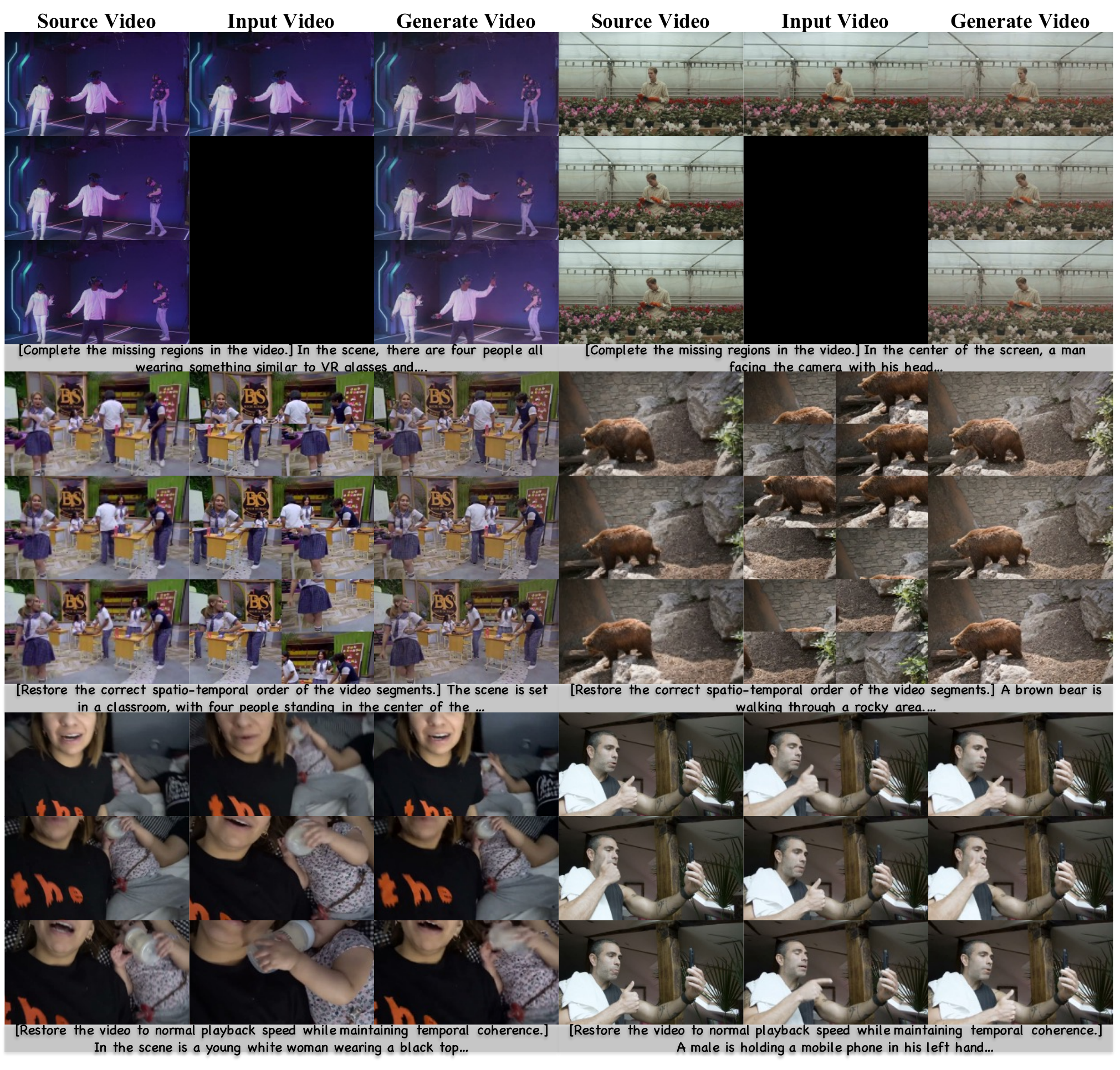} 
  \caption{\textbf{Illustration of Pretext Prediction.} }
  \label{fig:2}
\end{figure}

We visualize the model predictions for the three motion-centric pretext tasks used in Motion Alignment, including \textit{Cube Inpainting}, \textit{Speed Perturbation}, and \textit{Tube Shuffle}. As shown in Fig.~\ref{fig:2}, the model is able to (i) plausibly complete masked spatio-temporal regions, (ii) recover more natural motion dynamics from temporally perturbed inputs, and (iii) restore coherent spatio-temporal structure after tube permutation. These qualitative results indicate that the pretext objectives encourage the backbone to internalize motion cues and temporal reasoning, which benefits subsequent instruction-guided video editing.

\section{More qualitative results}
\label{sec:qualitative}

\begin{figure}[t]
  \centering
  \includegraphics[width=\textwidth,keepaspectratio]{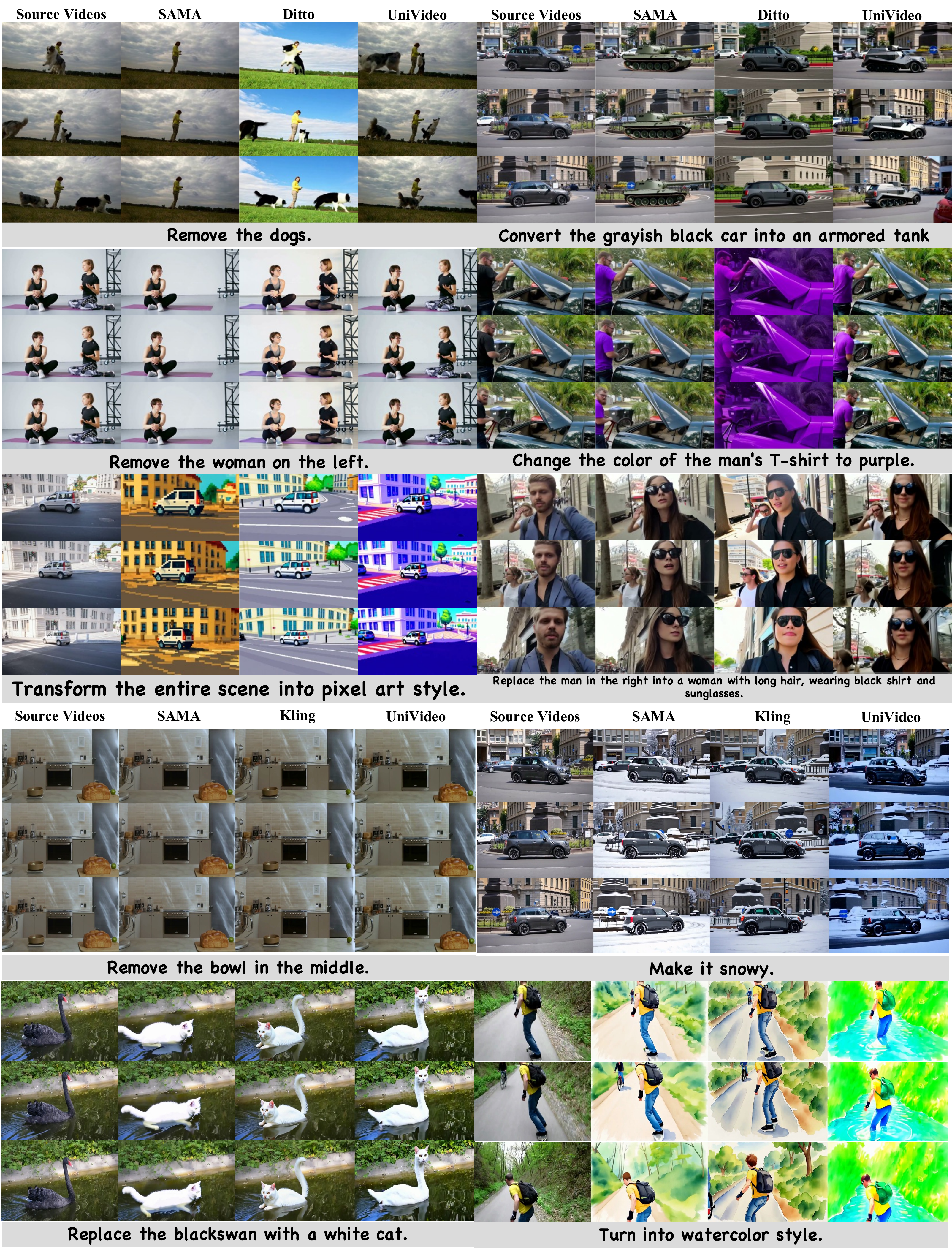}
  \caption{\textbf{More qualitative results on VIE-Bench}.}
  \label{fig:vie}
\end{figure}

\begin{figure}[t]
  \centering
  \includegraphics[width=\textwidth,keepaspectratio]{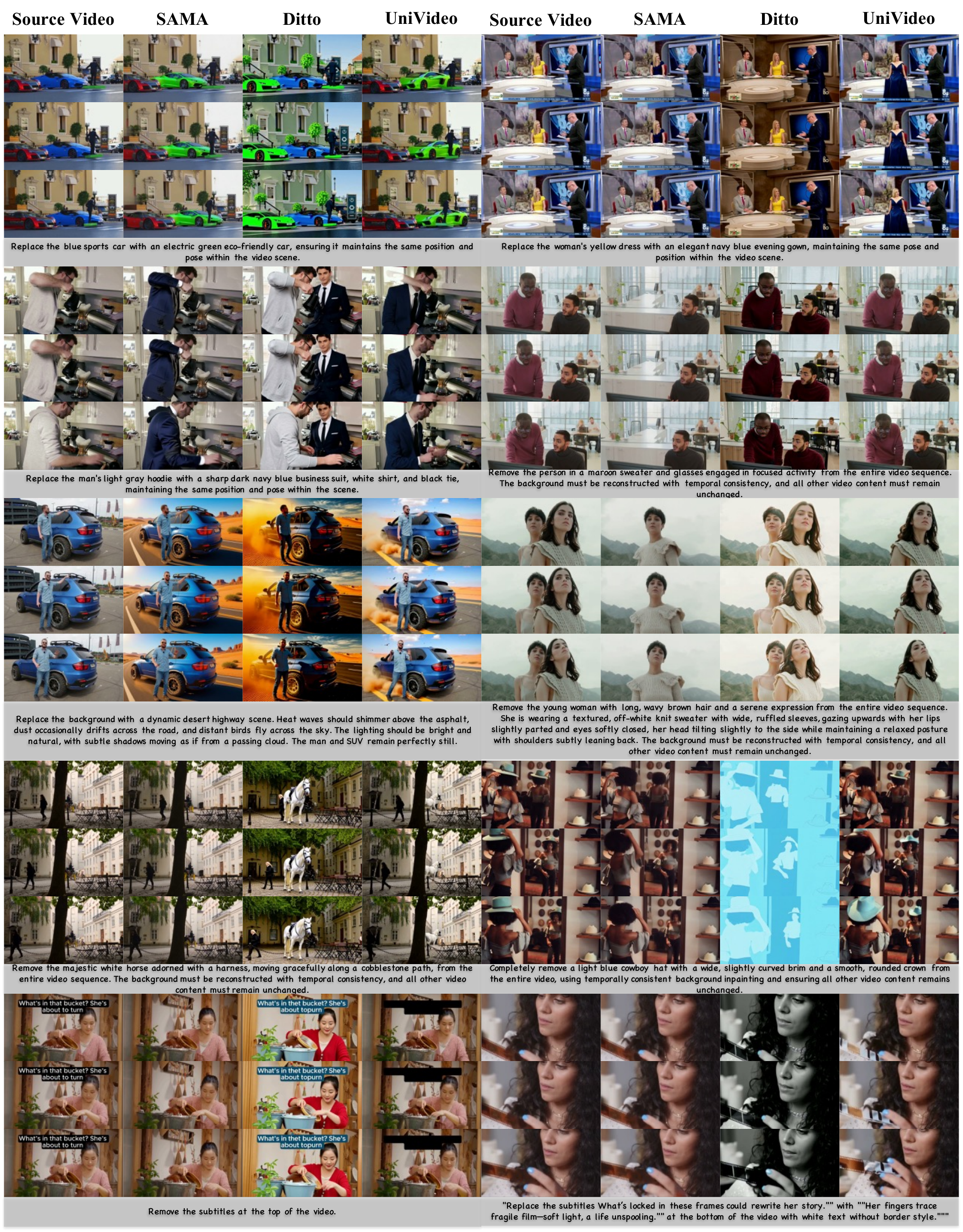}
  \caption{\textbf{More qualitative results on OpenVE-Bench}.}
  \label{fig:openve}
\end{figure}

\begin{figure}[t]
  \centering
  \includegraphics[width=\textwidth,keepaspectratio]{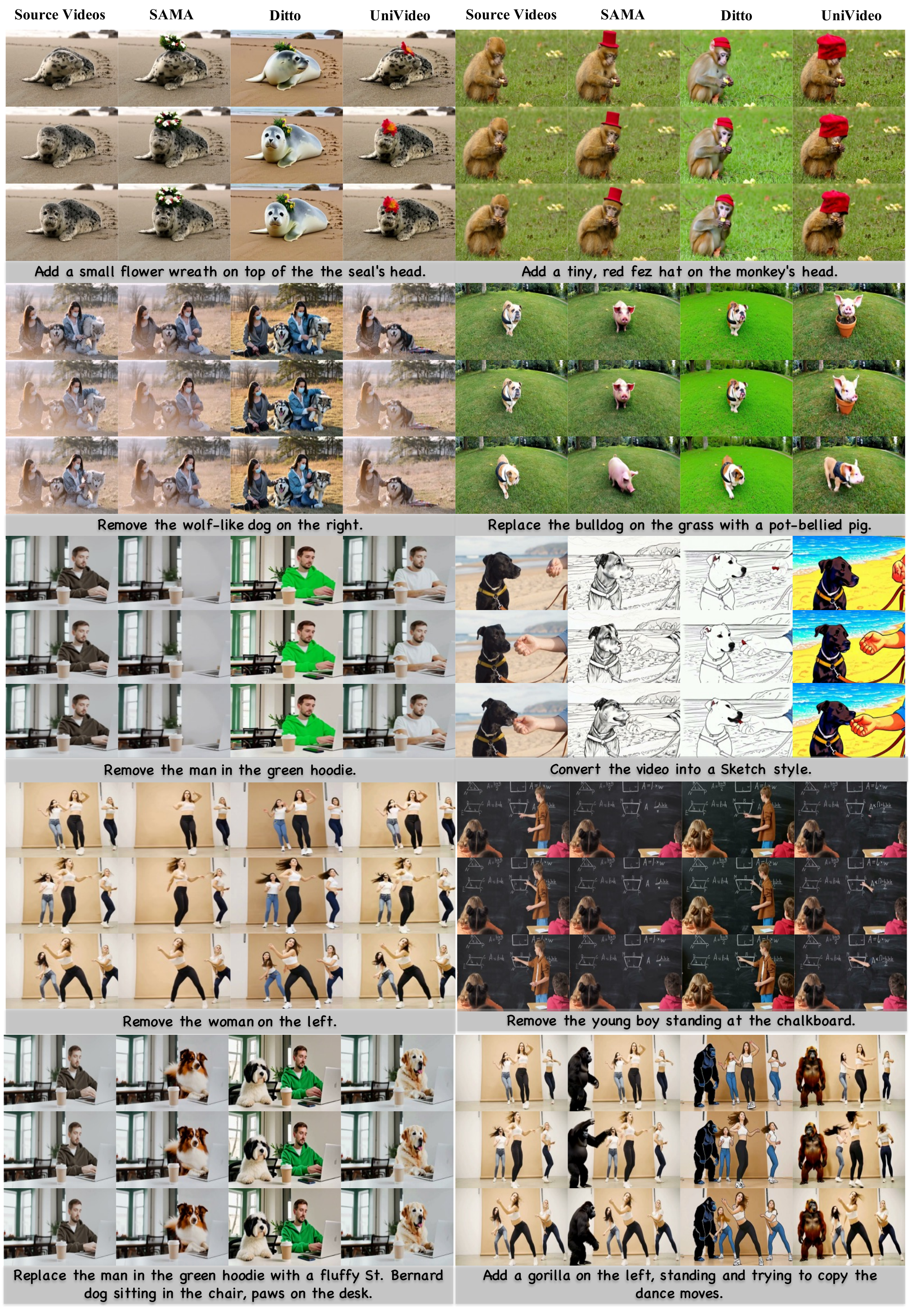}
  \caption{\textbf{More qualitative results on ReCo-Bench}.}
  \label{fig:reco}
\end{figure}

In this section, we present additional qualitative comparisons with other methods, showing that our method produces more consistent and visually appealing editing results across a wide range of scenarios.
More detailed visual comparisons are provided in Figs.~\ref{fig:vie}--\ref{fig:reco}.
The left column highlights examples where our method excels at semantic understanding and instruction grounding, while the right column presents cases emphasizing improved motion consistency and temporal alignment. 

